\pdfoutput=1

\documentclass[11pt]{article}

\usepackage[preprint]{acl}

\usepackage{times}
\usepackage{latexsym}

\usepackage{microtype}

\usepackage{inconsolata}

\usepackage{graphicx}

\usepackage[utf8]{inputenc} 
\usepackage[T1]{fontenc}    
\usepackage{hyperref}       
\usepackage{url}            
\usepackage{booktabs}       
\usepackage{amsfonts}       
\usepackage{nicefrac}       
\usepackage{microtype}      
\usepackage{xcolor}         
\usepackage{amsmath}
\usepackage{graphicx}
\usepackage{paralist}
\usepackage{wrapfig}
\usepackage{listings}

\usepackage{tikz}
\usepackage{float}
\usepackage{pgf}
\usepackage{pgfplots}

\pgfplotsset{compat=1.17}

\usepackage{colortbl}

\newcommand{\missingcitation}[1]{\textcolor{red}{[CITE]}}
\newcommand{\missingnumber}[1]{\textcolor{red}{[NUMBER]}}

\newcommand{\Sref}[1]{\S\ref{#1}}

\newcommand{\Fref}[1]{Figure~\ref{#1}}

\newcommand{\Aref}[1]{Appendix~\ref{#1}}

\usepackage{xspace}
\newcommand{\datasetname}{\textsc{ComPRed}\xspace}
\newcommand{\methodname}{\textsc{ComPO}\xspace}
\newcommand{\methodnamesmall}{\textsc{compo}\xspace}

\usepackage{enumitem}

\definecolor{codegreen}{rgb}{0,0.6,0}
\definecolor{codegray}{rgb}{0.5,0.5,0.5}
\definecolor{codepurple}{rgb}{0.58,0,0.82}
\definecolor{backcolour}{rgb}{0.95,0.95,0.92}

\newcommand{\applygradient}[1]{%
  \pgfmathparse{#1}%
  \let\val=\pgfmathresult 
  \pgfmathparse{int(\val)} 
  \let\intval=\pgfmathresult 
  \ifnum\intval<50
    \pgfmathsetmacro{\r}{255}
    \pgfmathsetmacro{\g}{128 * \val / 50}
  \else
    \pgfmathsetmacro{\r}{255 * (100 - \val) / 50}
    \pgfmathsetmacro{\g}{255}
  \fi
  \pgfmathtruncatemacro{\rint}{\r}%
  \pgfmathtruncatemacro{\gint}{\g}%
  \xdef\colorname{\noexpand\cellcolor[RGB]{\rint,\gint,0}}
  \colorname
  \val
}

\title{\methodname: Community Preferences for Language Model Personalization} 

\author{Sachin~Kumar$^{\clubsuit}$$\thanks{Equal contribution}$ \quad Chan~Young~Park$^{\heartsuit}$\footnotemark[1]  \\ \quad \textbf{Yulia~Tsvetkov}$^\heartsuit$ \quad \textbf{Noah~A.~Smith}$^{\heartsuit,\diamondsuit}$ \quad \textbf{Hannane~Hajishirzi}$^{\heartsuit, \diamondsuit}$ \\
$^\clubsuit$The Ohio State University, Columbus OH \\
$^\heartsuit$University of Washington, Seattle WA \\
$^\diamondsuit$Allen Institute for AI, Seattle WA \\
\texttt{\small kumar.1145@osu.edu, chanpark@cs.washington.edu}
}

\begin{document}

\maketitle

\begin{abstract}
Conventional algorithms for training language models (LMs) with human feedback rely on preferences that are assumed to account for an ``average'' user, disregarding subjectivity and finer-grained variations. 
Recent studies have raised concerns that aggregating such diverse and often contradictory human feedback to finetune models results in generic models that generate outputs not preferred by many users groups, as they tend to average out styles and norms.
To address this issue, we draw inspiration from recommendation systems,
and propose, \methodname, a method to personalize \textbf{p}reference \textbf{o}ptimization in LMs by contextualizing the probability distribution of model outputs with the \textit{preference provider}. 
Focusing on group-level preferences rather than individuals, we collect and release \datasetname, a question answering dataset with \textbf{com}munity-level \textbf{p}references from \textbf{Red}dit. This dataset facilitates studying diversity in preferences without incurring privacy concerns associated with individual feedback.
Our experiments reveal that  conditioning language models on a community identifier (i.e., subreddit name) during preference tuning
substantially enhances model performance. Conversely, replacing this context with random subreddit identifiers significantly diminishes performance, highlighting the effectiveness of our approach in tailoring responses to communities' preferences.\footnote{Data: \url{https://huggingface.co/datasets/allenai/compred}, Code: \url{https://github.com/allenai/compred}} 
\end{abstract}

\section{Introduction}
Language models have become ubiquitous in user-facing applications, offering an opportunity for refinement through user feedback~\citep{ouyang2022training}. A common approach to this refinement process is preference tuning, where users indicate their preference between two model-generated outputs, which is then used to adjust the model weights with the aim of generating outputs generally preferred by humans~\citep{christiano2017deep,rafailov2024direct}. However, annotating preferences is inherently \textit{subjective}~\citep{kirk-etal-2023-past}.
Current methodologies typically aggregate these (often contradicting) preferences without considering individual differences, resulting in models tuned towards a hypothetical ``average'' user~\citep{bakker2022fine,chakraborty2024maxmin}. 

Previous research has explored different methods to address this issue. Some approaches focus on value pluralism~\citep{sorensen2024value,sorensen2024roadmap,bakker2022fine}---aiming to present balanced views for inputs where different users may hold different values, such as subjective or opinion-based questions. 
Work on system instructions~\cite{achiam2023gpt} allows users to verbalize their preferences as part of the input, such as preferred content and style. 
Additionally, there have been efforts to categorize human feedback into underlying factors such as demographics, culture, and stylistic preferences that advocate for factor-specific fine-tuning~\citep{jang2023personalized}. 
However, preferences can be nuanced and often implicit. Verbalizing them or factorizing them into interpretable dimensions, however, may not always be feasible or practical in user-facing applications~\citep{cunningham2022implicit}.

In contrast to prior work which tends to separate the preference provider from the preferences, in this work, we propose to model the provider directly. 
Inspired by research in recommender systems~\citep{10.1145/3285029}, we propose to contextualize LMs with information about the users providing the preferences. That is, instead of $p(\mathbf{y} \mid \mathbf{x})$ where $\mathbf{x}$ and $\mathbf{y}$ are user input and model output respectively, we propose to model $p(\mathbf{y} \mid \mathbf{x}, \text{user})$ where $\text{user}$ provides the preference. Given a preference dataset marked with this additional information, we make simple modifications to the supervised finetuning and preference tuning stages to incorporate this context (\Sref{sec:method}).  

Since gathering extensive user-specific preference data is infeasible for academic research, we validate our methodology by training models personalized to Reddit communities (i.e., subreddits). Reddit's post-comment format naturally supports building question-answering models where posts can serve as model inputs and comments as outputs, respectively. Additionally, subreddit users vote on others' comments, allowing us to use collective upvotes as a proxy for community preference. Reddit consists of thousands of subreddits often discussing similar topics but different participants and different upvoting patterns. As a result, this data offers a readily available resource of diverse preferences to learn from. 
We refer to this approach as \textbf{Com}munity \textbf{P}reference \textbf{O}ptimization (\methodname).

Our aim is to study whether incorporating the community's context during preference tuning results in models generating responses tailored towards the communities' preferences. 
To that end, we collect and release \datasetname, a preference dataset comprising five different groups of subreddit (covering themes related to gender, politics, history, science, and finance). Each group constitutes subreddits that discuss similar topics but differ from each other in values or norms and thus diverge in preferences (e.g., \texttt{r/askliberal} vs.~\texttt{r/conservative} under politics). 
With experiments using direct preference optimization (DPO; ~\citealp{rafailov2024direct}), we find that, across all domains, adding subreddit context indeed leads to more tailored responses, which are preferred both by human annotators and automated metrics. Conversely, conditioning on the wrong subreddit yields inferior results, indicating that generating outputs preferred by a different user is detrimental to the current user's experience. Ultimately, our work introduces \textbf{a novel dataset} and \textbf{method to personalize LMs to fine-grained communities}, 
paving the way towards personalized and adaptive AI assistants.

\begin{figure*}[t]
    \centering
    \includegraphics[width=\textwidth]{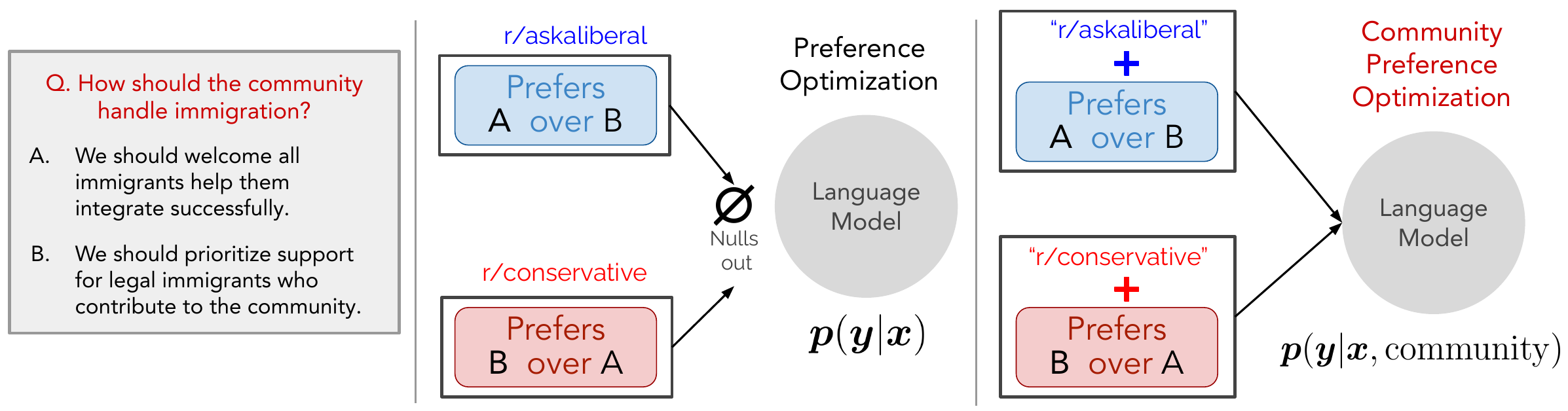}
    \caption{
    \textbf{Conceptual overview of community preference optimization}: 
    When asked about immigration, two communities may prefer different answers (A and B). Conventional preference optimization aggregates these conflicting responses, often averaging them out or reflecting the majority view. Our proposed community preference optimization incorporates subreddit-specific contexts into the model, tailoring outputs to align with the distinct norms and values of individual communities.
    }
    \label{fig:figure1}
\end{figure*}

\section{Related Work}
\paragraph{Training LMs to align with human preferences} While human annotations have been used in NLP for decades, aligning LMs with human feedback (RLHF) marked a key paradigm shift in the field~\citep{ouyang2022training, rafailov2024direct}. Yet popular alignment techniques ignore the subjectivity inherent in human preferences \citep{davani2022dealing,kirk-etal-2023-past,kirk2024prism}. Aggregating such preferences risks dilution leading to models optimized towards a ``generic'' human that may not satisfy any particular user group well \citep{kirk2024benefits}. 
Prior work in addressing these issues includes the investigation of value pluralism 
 \citep{sorensen2024roadmap} that attempts to summarize different perspectives, particularly in contexts where users' values diverge~\citep{bakker2022fine}.
 Another line of work has focused on factorizing subjectivity into explainable dimensions (such as style, complexity, among others)~\citep{jang2023personalized} or proposed that users can personalize their models by simply verbalizing their preferences in the instructions, such as desired content and style~\citep{achiam2023gpt}. It is, however, not always feasible or practical for users to articulate their preferences~\citep{cunningham2022implicit}. 
Most closely related to our work is research on incorporating social and cultural context in NLP systems \citep{solaiman2021process, qiu2022valuenet,bang-etal-2023-enabling,kumar2024genz}. \citet{li2024personalized} in particular proposes a similar paradigm to ours but their experiments are limited to summarization tasks with a small set of annotators, while we conduct large-scale experiments on a broader range of communities.

\paragraph{Personalization in Recommender Systems}
Our work draws parallels to recommender systems where personalization serves to match the right types of services, products, or content to the right users intended to improve user engagement whereas showing every user the same recommendation is undesirable. Collaborative filtering, a widely adopted technique in recommender systems,  matches different users with similar tastes~\citep{koren2021advances}. While early work in this space relied on learning linear separable or interpretable features for each user and product~\citep{rendle2010factorization}, those approaches have been surpassed by neural network-based methods to learn user and product embeddings as a means to capture the underlying preferences and
intrinsic characteristics of users and items~\citep{mnih2007probabilistic,zhao2023embedding}. We draw inspiration from this line of work to model preference providers directly in language models.

\paragraph{Preference Tuning Datasets}
Quality datasets have been the primary driver of  advances in NLP. While several human-annotated preference datasets are publicly available~\citep[][\textit{inter alia}]{bai2022training,nakano2021webgpt,kaufmann2023survey}, they have been primarily focused on aspects such as helpfulness and harmlessness and are collected with a goal of obtaining high agreement among annotators or between researchers and annotators~\citep{ziegler2019fine,stiennon2020learning} rather than embracing subjectivity. Our work is related to~\citet{jang2023personalized}, who collect a synthetic data of factor-based preferences using GPT-4 as a judge; we rely on natural existing community-based preferences. Most closely related to our work is~\citet[][SHP]{pmlr-v162-ethayarajh22a} who also collect a Reddit based preference dataset. They largely
focus on knowledge-seeking subreddits aimed at improving general purpose alignment. Our goal  is modeling diversity in preferences; hence, we curate a different set of subreddits where
divergences are more prevalent.

\section{Community Preference Optimization}
\label{sec:method}
To build conversational or instruction following language models $p_\theta(\mathbf{y} | \mathbf{x})$, a typical pipeline starts with pretraining, followed by supervised 
finetuning (SFT) to obtain $p_\mathrm{sft}$ and finally preference tuning. SFT requires a dataset with instances of the form $(\mathbf{x}, \mathbf{y})$ where $\mathbf{x}$ is the input query in natural language, and $\mathbf{y}$ is the expected output. 
Preference tuning is done with a dataset comprising prompts, preferred answers, and dispreferred answers, $(\mathbf{x}, \mathbf{y}_j, \mathbf{y}_k)$.
While the standard recipe for preference tuning relies on reinforcement learning (RLHF), in recent work, \citet{rafailov2024direct} proposed direct preference optimization (DPO), which implicitly optimizes the same objective as RLHF but 
offers higher stability. Hence, we adopt this framework for all experiments in this work.\footnote{Our goal in this work is to understand the effect of personalization, which is orthogonal to the choice of the preference tuning approach. Our method can, in principle, be applied to any preference tuning approach. 
} DPO's training objective is as follows
\begin{align*}
\resizebox{\columnwidth}{!}{
$-\mathbb{E}_{(\mathbf{x}, \mathbf{y}_j, \mathbf{y}_k)} \left[ \log \sigma \left( \beta \log \textstyle\frac{p_\theta (\mathbf{y}_j \mid \mathbf{x})}{p_\mathrm{sft} (\mathbf{y}_j \mid \mathbf{x})} - \beta \log \textstyle\frac{p_\theta (\mathbf{y}_k \mid \mathbf{x})} {p_\mathrm{sft} (\mathbf{y}_k \mid \mathbf{x})} \right)   \right]$
}
\label{eq:dpo}
\end{align*}
where $\sigma$ is the sigmoid function and $\beta$ is a hyperparameter. The loss is optimized with respect to $\theta$ where $p_\theta$ is initialized with $p_\mathrm{sft}$, which is often referred to as the reference model, but the latter is kept frozen during this stage.  

Our goal is to build a community-personalized model, $p_\theta(\mathbf{y} \mid \mathbf{x}, r)$, where $r$ is the subreddit to which the model will be personalized towards. To train this model, we collect a preference dataset from Reddit along with the subreddit:  tuples of the form $(r, \mathbf{x}, \mathbf{y}_j, \mathbf{y}_k)$. The subreddit $r$ is  represented by its name (such as \texttt{r/conservative}).\footnote{While special tokens for each $r$ may also be considered for this task, prior work in personalization in NLP has shown that choosing tokens from the existing vocabulary can perform just as well without blowing up the vocabulary size when the number of users increases~\citep{mireshghallah2021useridentifier}.} The same pair of answers ($\mathbf{y}_j$,  $\mathbf{y}_k$) may have different preferences depending on the subreddit $r$ (see \autoref{fig:figure1}). To incorporate $r$ into the pipeline, we  concatenate the subreddit name in front of the input text $\mathbf{x}$ for both the SFT and DPO stages. We refer to this approach as community preference optimization. In what follows, we detail our data collection strategy.

\section{\datasetname: A Dataset of Diverse Community Preferences from Reddit}
\label{sec:data}

The Reddit platform is divided into community forums known as subreddits. The majority of subreddits are created based on discussion topics; however, new subreddits are often formed where a subset of the community diverges from the existing ones. These divergences stem from multiple factors, including topic specialization, user demographic differences, value disparities, and different community norms~\citep{Hessel2016ScienceAA,10.1145/3512908}. As a result, the preferences of the communities for what discussions they encourage are different. For instance, as illustrated in \autoref{fig:figure1}, the subreddits \texttt{r/askliberal} and \texttt{r/Conservative} have  different values and the same two comments posted in both forums to the same post will receive vastly different upvotes. Note that while some subreddits explicitly encode their rules and norms, these preferences are implicit and more nuanced; the members generally do not verbalize their reasons for upvoting or downvoting a comment. This setup mirrors a chat-based LM where a user may not be able to or do not explicitly verbalize their preferences. 

Our work capitalizes on these factors to collect diverging preference datasets across different subreddits. Each instance in our dataset consists of the subreddit name, a question, a preferred answer, and a dispreferred answer. We begin with a collection of threads. Each thread consists of a main post $q$ posted in the subreddit $r$ along with $N\geq0$ comments made by other users: $\{(a_1, u_1, t_1), \ldots, (a_N, u_N, t_N)\}$ where $a_i$ denotes the comment text posted at time $t_i$ with $u_i$ being the number of upvotes it receives.\footnote{We only look at first-level comments to the original post and discard the remaining comment tree. Also, we discard user identifiers of original posters and the commenters.} For each pair of comments $a_i$ and $a_j$ in this thread, we add to our dataset a preference instance $(r, q, a_i, a_j)$ if $t_i > t_j$ and $u_i > u_j$. That is, the comment $a_i$ is preferred over $a_j$ if it was posted after $a_j$ and still received more upvotes than the latter. We adopt this strategy following the \href{https://huggingface.co/datasets/stanfordnlp/SHP}{Stanford Human Preference dataset}~\citep{pmlr-v162-ethayarajh22a} which argues that using the post time along with the upvote counts provides a more robust signal of preference than just using upvotes as it avoids recency bias. Early comments receive higher upvotes on average due to more exposure time, but if a comment has received more upvotes despite being posted later, it is a clearer signal of preference.

Notably, preferences are conditional on the subreddit in which the post was made. This means that answering the same prompts with same answers may elicit different preferences in distinct subreddits. Our approach aims to capture general trends in preferences rather than specific instances, aligning with real-world scenarios where users provide feedback on model-generated responses. 



\subsection{Details of Data Collection}


We create \datasetname (\textbf{Com}munity \textbf{P}references from \textbf{Red}dit) using the Reddit subset of Dolma~\citep{soldaini2024dolma} which was collected using Pushshift API~\citep{baumgartner2020pushshift} in April 2023.\footnote{This is the last licensed publicly available Reddit dataset as the Pushshift API is no longer supported.} We manually select subreddits from this collection to study divergences along different dimensions categorized in five sets grouped by their general topics and themes:

\begin{itemize}[leftmargin=3mm,topsep=0mm,itemsep=0mm]
\item {\bf Science:} consisting of $71$ subreddits 
 discussing various scientific disciplines. With this set, we aim to study divergences arising from topic specialization. For example, the same question asked in \texttt{/r/science} and \texttt{/r/StringTheory} may get different details in the answers.
\item {\bf Finance:} consisting of $11$ subreddits focused on the discussion of topics related to finance, investing, and money. While often containing similar questions, subreddits in this set are divided based on income brackets and differing investing goals. For example, while \texttt{/r/personalfinance} is for general finance management advice for middle-income individuals, \texttt{r/financeindependence}
is focused on early retirement. 
\item {\bf History:} consisting on $5$ subreddits related to history. This set consists of a mix of topic specialized communities as well as those that have different norms. For example, \texttt{/r/AskHistorians} and \texttt{/r/History} contain similar questions but the former expects detailed academic answers with cited sources whereas the latter does not enforce such rules.
\item {\bf Politics:} consisting on $63$ subreddits related to politics divided based on political leanings (\texttt{r/Conservative}), political issues (\texttt{r/gunpolitics}), countries (\texttt{r/ukpolitics}), politicians (\texttt{r/sandersforpresident}), \textit{inter alia}. This collection is aimed at studying changing preferences with different community values.
\item {\bf Gender/Sexuality:} consisting on $37$ subreddits related to gender and sexuality often intersected with other personal attributes and topics such as occupation, fitness, fashion, parenting, etc. With this collection, we seek to understand personalization effects with demographics. 
\end{itemize}

\begin{table}[t]
\centering
\resizebox{\columnwidth}{!}{
\begin{tabular}{lcccccc}
\textbf{Domain} & Finance & Science & History & Politics & Gender & Total \\ \hline  \textbf{Train} & 232,306 & 160,854 & 28,645 & 242,485 & 706,981 & 1,371,271 \\ 
\textbf{Test} 
  & 2,780 & 3,013  & 591 & 2,937  & 9,158 & 18,479 \\ \hline
\end{tabular}
}
\caption{Number of train and test examples in each domain in our \textsc{ComPRed} dataset. The train examples indicate preference pairs while the test examples are prompts only.
}
\label{tab:overall_data_statistics}
\end{table}


To each thread, we apply several quality filters. We remove any instances where  
\begin{inparaenum}[(i)]
\item the post or the comments are not in English using the FastText language classifier~\citep{joulin2016bag};
\item the post or the comments contain non-textual content like images, video, URLs, mentions of Reddit users (\texttt{/u/<username>}), or the word Reddit itself;
\item the post or the comments contain adult content (using the field \texttt{over\_18});
\item the post or the comments have been pinned, or stickied; 
\item one or both comments have been posted by the original poster or been deleted or updated since being originally posted;
\item the two comments have a length difference larger than a threshold (we measure it using the ratio of number of sentences in each comment and discard instances where the ratio is greater than 5); or
\item the post is not a question. We use an off-the-shelf \href{https://huggingface.co/shahrukhx01/bert-mini-finetune-question-detection}{question identifier model} 
from Huggingface for this task.
\end{inparaenum}
First, we randomly subsample $2.5\%$ of the threads whose posts are used for final evaluation (this may include posts with single comments). We convert the remaining threads to preference instances using the method described earlier. 
Further, following SHP~\citep{pmlr-v162-ethayarajh22a}, for each thread, we filter out instances where the upvote ratio of the preferred to the dispreferred answer is less than $2.0$. Finally, to limit the number of instance from each thread, we randomly subsample at most 5 instances from each thread. 
For the evaluation set, while we do not use the comments for the final evaluation, we do not discard them since they can be useful for evaluating reward models trained on this dataset (see \Sref{sec:reward-eval}).
In total, we collect over 1M preference pairs across 187 subreddits. The statistics of the datasets are provided in \autoref{tab:overall_data_statistics} and breakdowns per subreddit provided in \Aref{appsec:data_stats}.


\section{Motivational Experiment: Context Aware Reward Training}
\label{sec:reward-eval}


The driving motivation and the primary hypothesis in this work is that subreddit context aware models should generate answers that the respective community will prefer over the ones that are not context-aware. 
To solidify this motivation, 
we first train reward models using the preference data with and without providing the model with community context and measure changes in their preference prediction accuracy. Although we do not use the reward models directly in our model training experiments, prior work has shown that reward accuracy is a good indicator of preference data quality~\citep{lambert2024rewardbench}. We train the models using a binary classification objective following the Bradley-Terry approach~\citep{bradley1952rank} as used by prior work~\citep{ouyang2022training}.
\begin{align*}
    \mathcal{L}_\mathrm{reward} = - \log \sigma (f_\theta(r, \mathbf{x}, \mathbf{y}_j) - f_\theta(r, \mathbf{x}, \mathbf{y}_k)) 
\end{align*}
Here, $f_\theta$ is a parameterized reward function (Llama 2 7B~\citep{touvron2023llama} with a classification head in our case) which takes as input the concatenation of the subreddit name $r$, question $\mathbf{x}$, and a response $\mathbf{y}$ and predicts a scalar. Training $f_\theta$ with this loss leads to the preferred response getting a higher reward than the dispreferred one. In this experiment, we use this model as binary classifier to measure classification accuracy for our held out evaluation set, i.e., we compute, given a set of instances containing the preferred and the dispreferred responses, what fraction of them get a higher score for the preferred one. For comparison, we train another version the reward model where the subreddit information is not provided (no-context). 
\begin{table}
\resizebox{\columnwidth}{!}{
\begin{tabular}{@{}llllll@{}}
\toprule
 & \textbf{Science} & \textbf{History} & \textbf{Finance} & \textbf{Gender} & \textbf{Politics} \\ \midrule
\textbf{no-context} & 66.1  & 64.1 & 65.7 & 63.3  & 63.4 \\
\textbf{context} & 66.2 & \textbf{65.2} & 66.0 & \textbf{64.5}  & \textbf{64.2} \\ \bottomrule
\end{tabular}
}
\caption{Overall preference accuracy of reward models trained with and without the subreddit context on the comment pairs on the test sets. Subreddit-specific performance differences can be found in \Aref{appsec:subreddit-reward-accuracies}. Datasets where the improvement >1\% are \textbf{bolded}.
}
\label{tab:reward-accuracies}
\end{table}

We report the average accuracies for each of our datasets in \autoref{tab:reward-accuracies} and subreddit-wise accuracies in \Aref{appsec:subreddit-reward-accuracies}. Overall, adding subreddit information leads to higher prediction accuracies, suggesting the importance of providing preference as additional context. The improvements are more pronounced in datasets such as gender and politics where we expect higher dissent amongst different communities due to differences in values. For datasets where the overall improvements are minimal, a closer look at subreddit specific performance still reveals that the context is helpful in majority of the subreddits (\Aref{appsec:subreddit-prediction-accuracies}). For those, where we do not observe improvements, we hypothesize that the context is redundant (see \Sref{sec:analysis} for more details). It is noteworthy that this experiment is meant to motivate the case for personalization in RLHF. We do not actually use these reward models in our main experiment that follows since we conduct experiments with DPO \citep{rafailov2024direct}. 

\section{Main Experiments: \methodname}
We first describe our experimental setup and baselines. Next, we present results of large-scale automatic evaluations followed by human evaluation.
At the end, we present analysis studying the impact of training dataset size and subreddit predictibality on the degree of personalization.

\paragraph{Implementation Details}
We use Llama-2 7B as the base model \cite{touvron2023llama} and conduct all finetuning experiments using low-rank adapters (LoRA; \citealp{hu2022lora}) to keep memory requirements low. For each of the datasets, we first conduct SFT 
on this model to generate preferred answers conditioned on the subreddit name and the question to obtain $p_\mathrm{sft}$. We then continue training the LoRA weights using the DPO objective 
to obtain our final models. Training stage hyperparameters are provided in \Aref{appsec:hyperparameters}. 
We generate outputs for the evaluation sets using top-$p$ sampling~\citep{Holtzman2020The} with $p=0.95$ with a temperature of $0.7$ following \citet{zephyr}. 

\paragraph{Baselines}
We seek to understand the benefits of \textbf{c}ommunity-contextualized preference tuning (we refer to our final model as \methodname). 
Hence, we compare this model with baselines that are \textbf{n}on-\textbf{c}ontextualized, are contextualized but not preference tuned, or neither. Our baselines are

\noindent{\bf \textsc{sft-nc}}, where we finetune the base model only on (question, preferred answer).

\noindent{\bf \textsc{dpo}}, where we preference tune \textsc{sft-nc} but without providing the subreddit information.

\noindent{\bf \textsc{sft-c}}, where we finetune the base model only on (question, preferred answer) contextualized with the subreddit identifier.

Note that we do not specify the target subreddit for other baselines except for \textsc{sft-c}. In our initial exploration, including the target subreddit in other baselines did not lead to improvements and, in some cases, resulted in disfluent outputs for certain subreddits. Given the resource and cost intensity of GPT-4 evaluations, we decided not to include this version as a baseline.
Additionally, as we discuss in \Sref{sec:data}, we aim to isolate and study nuanced implicit preferences expressed in upvote patterns which are not verbalized or factorized along explainable dimensions. Hence, existing methods such as factor-based personalization approaches \citep{jang2023personalized} or simply providing preferences in the (system) prompt are not applicable to our setup.\footnote{While certain subreddits contain descriptions prescribing their norms and values, they often do not capture all nuances. In any case, most subreddits in our dataset do not contain any descriptions.}
\subsection{Automatic Evaluation}

Our models are essentially trained to generate Reddit comments preferred by the relevant community, or comments that should get high upvotes. 
To evaluate this, first, we measure if adding the community context results in better responses as compared to responses without this context. Second, we evaluate if changing the community context to a random subreddit degrades the generated answer. 

Since a large number of our questions are subjective, they do not have a reference answer. Thus, we adopt a reference-free metric. We evaluate using GPT-4 as a judge~\citep{achiam2023gpt},\footnote{We use version \texttt{gpt-4-1106-preview}.} which has been shown to correlate well with human judgments~\citep{zheng2024judging}.
Specifically, we present the subreddit information, the question, and two answers to the judge, and ask it to select the one which will get more upvotes if posted as an answer in the respective subreddit (exact prompts in \autoref{fig:gpt4-prompts} in \Aref{appsec:hyperparameters}). Our goal is not to evaluate the absolute quality of an answer but to evaluate the relevant community's preference. 
We follow an A/B testing framework where we randomize the order of the answers presented to the model to prevent any positional biases. We report win-rate as our final metric which when comparing models $m_1$ and $m_2$ measures the fraction of $m_2$ generated responses that were selected by the judge. The overall results are summarized in \autoref{fig:all-win-rates} 
and \autoref{tab:dpo-randomized-subreddits}. We provide granular win rates based on individual subreddits in \Aref{appsec:subreddit-win-rates}.

\begin{table*}[htbp]
\resizebox{\linewidth}{!}{
\centering
\begin{tabular}{cccccccccccc}
 & \textsc{sft-c} & \textsc{dpo} & \methodnamesmall &  & \textsc{sft-c} & \textsc{dpo} & \methodnamesmall &  & \textsc{sft-c} & \textsc{dpo} & \methodnamesmall \\
\textsc{sft-nc} & \applygradient{49.64} & \applygradient{56.06} & \applygradient{56.97} &  & \applygradient{49.95} & \applygradient{55.31} & \applygradient{56.20} &   & \applygradient{55.16} & \applygradient{62.86} & \applygradient{64.52} \\
\textsc{sft-c} &  & \applygradient{58.16} & \applygradient{58.40} &  &  & \applygradient{57.41} & \applygradient{57.40} &  &  & \applygradient{65.36} & \applygradient{64.77} \\
\textsc{dpo} &  &  & \applygradient{56.67} &  &  &  & \applygradient{55.64} &  &  &  & \applygradient{62.27} \\
 & \multicolumn{3}{c}{Finance} &  & \multicolumn{3}{c}{Science} &  & \multicolumn{3}{c}{History} \\
\textsc{sft-nc} & \applygradient{50.38} & \applygradient{58.18} & \applygradient{59.93} &  & \applygradient{52.17} & \applygradient{55.05} & \applygradient{55.56} &  &  &  &  \\
\textsc{sft-c} &  & \applygradient{61.42} & \applygradient{61.95} &  &  & \applygradient{56.16} & \applygradient{56.16} &  &  &  &  \\
\textsc{dpo} &  &  & \applygradient{59.69} &  &  &  & \applygradient{54.98} &  &  &  &  \\
 & \multicolumn{3}{c}{Politics} &  & \multicolumn{3}{c}{Gender/Sexuality} &  &  &  & 
\end{tabular}
}
\caption{Aggregated win-rates for each dataset in \datasetname. For each cell, the win-rate is computed as the percentage of examples for which the model specified in the column is preferred over the one specified in the row. 
}
\label{fig:all-win-rates}
\end{table*}

\begin{table}
\centering
\resizebox{\columnwidth}{!}{
\begin{tabular}{lccccc}
\toprule
& \textbf{Finance} & \textbf{Science} & \textbf{History} & \textbf{Politics} & \textbf{Gender \& Sexuality} \\
\midrule
\textbf{Win-rate} & 47.77 & 44.23 & 43.82 & 41.19 & 46.05 \\
\bottomrule
\end{tabular}
}
\caption{
 Aggregated win-rates of \textsc{compo-randomized} versus \methodnamesmall. Responses generated by conditioning on the wrong community are less preferred by the judge.}
\label{tab:dpo-randomized-subreddits}
\end{table}

\paragraph{Do the models learn to rely on community context?}
Comparing the SFT models (\textsc{sft-c} over \textsc{sft-nc}), we do not see clear signal of whether context is helpful as all win-rates lie close to $50\%$ suggesting that \textbf{supervised finetuning alone does not result in personalization}. 
Indeed, \textsc{dpo} wins over both SFT models, highlighting the importance of preference tuning. Our proposed approach \methodname wins over
all the baselines including \textsc{dpo}, highlighting the importance of personalization. The improvements are most pronounced in political and history related subreddits. 

\paragraph{Do the models learn to rely on the \textit{right} community context?}
The previous experiment established that contextualizing on community information can be beneficial than not having any context. In this experiment, we test the importance of contextualizing on the right community. Using the \methodname model, we generate responses for our evaluation sets but switch the community context to a random subreddit sampled from the list of all subreddits from the respective datasets (we call this method \textsc{compo-randomized}). As shown in \autoref{tab:dpo-randomized-subreddits}, across all datasets, random context leads to a decline in performance providing evidence for the importance of the right context.  


\subsection{Human Evaluation}
Having observed positive signals from automatic evaluations, we turn to more reliable human judgments.

\noindent \textbf{Annotator Selection}
To compare our \methodname model to our best baseline, \textsc{DPO}, we ideally would have asked the community members themselves. However, recruiting enough annotators from all the subreddits we studied proved challenging.
Instead, we focused on finding participants who closely matched the actual members of each subreddit. For example, to evaluate \texttt{r/askwomen}, we recruited annotators who identified as women and were active on Reddit (self-reported on \href{https://prolific.com/}{Prolific}). Additionally, before starting their annotations, we asked each annotator to rate their familiarity with the subreddit they were evaluating (new, occasional, or active user).
Of our annotators, 58\% were at least occasional users, and 13\% were daily users. For those who were not already part of the community, we asked them to familiarize themselves by browsing multiple posts and comments, focusing on upvote counts, for 5-10 minutes.

\noindent \textbf{Annotation Task and Process} 
Given a question and two model responses, annotators were asked to identify which response was more likely to receive more upvotes in the subreddit under consideration, with an option of a tie. The order of the two responses was randomized during the annotation process. Annotators could also flag if one or both models generated gibberish or repetitive outputs.

Each subreddit had at least three annotators (when we have 3+ annotators, we chose the top three annotators with the highest agreement), and we annotated 30 examples per subreddit (except for \texttt{r/conservative} where we annotated 29 samples). We evaluated 8 subreddits across four domains (finance, history, gender, and politics)\footnote{We chose not to conduct human evaluation for the science evaluation set, as they required specialized domain expertise, making the tasks much harder and annotations unreliable.}, resulting in a total of 717 human annotations.
The average annotator agreement of preference labels, measured by Fleiss's $\kappa$, was 0.36, which is reasonable \citep{bakker2022fine} given the difficulty and subjectivity of the task. Human annotators' majority labels agreed with GPT-4 evaluation results 69.8\% of the time.
The user interface for human annotation, full guidelines, annotator pay, and additional details can be found in \Aref{appendix:human-annotation}.

\begin{figure}[t]
    \vspace{-6pt}
    \centering
    \includegraphics[width=0.5\textwidth]{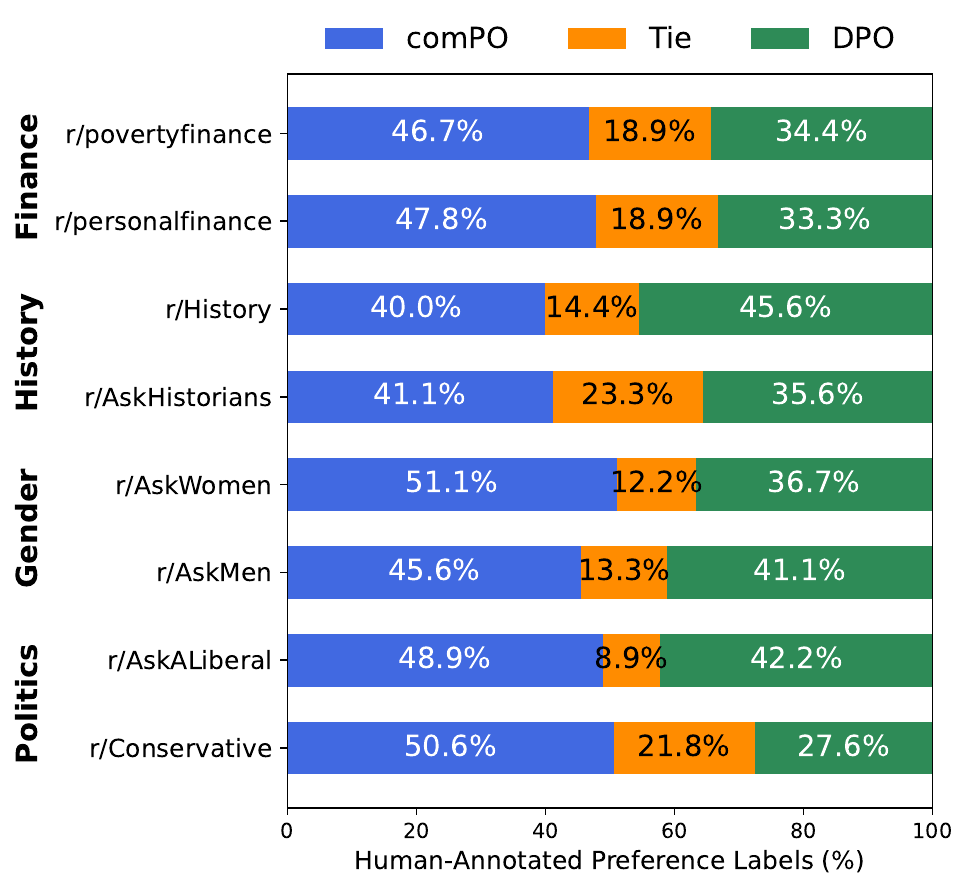}
    \vspace{-21pt}
    \caption{{\bf Human Evaluation.} The proportions of human annotators' preference labels for our model (\methodname) and the baseline (\textsc{dpo-nc}). 
    } 
    \label{fig:human-annotation}
\end{figure}

\noindent \textbf{Annotation Results}
On average, the percentage of responses that any of the three annotators marked as gibberish was 14.2\% for \methodname and 18.8\% for \textsc{DPO}. 
\Fref{fig:human-annotation} presents the proportion of preference labels by human annotators. The results show that responses generated by \methodname are favored on average more than those from the baselines (46.5\% vs.~37.1\%). 
Our model performs particularly well in politics and gender, especially in \texttt{r/Conservative} (50.6\% vs.~27.6\%) and \texttt{r/AskWomen} (51.1\% vs.~36.7\%).
Previous research has demonstrated that LM outputs are often biased, exhibiting a tendency to be relatively more liberal \citep{feng-etal-2023-pretraining, bang2024measuring} and to reflect men's voices more prominently \citep{wong2023chatgpt}. Our results suggest that providing community context for less dominant voices can help alleviate this issue.

For history related subreddits, we obtain mixed results contrasting with automatic evaluation. We speculate that limited improvements on these subreddits could be due to the smaller size of the history dataset compared to others or a relatively small size of our base model (7B). Future work on training larger models may show a clearer signal but our computation budget prevents us from conducting such experiments.

\subsection{Analysis}
\label{sec:analysis}


Our evaluation shows that while our proposed approach performs better than the baselines, its win-rates are far from being $100\%$, indicating that subreddit context is not \textit{always} useful. We hypothesize that if the subreddit can be predicted from the input text $\mathbf{x}$, contextualization becomes redundant (and thus does not help). We operationalize the \textit{predictability} of the subreddit using a classification model which takes as input $\mathbf{x}$ and is trained to predict the subreddit in which it was posted (details in \Aref{appsec:subreddit-prediction-accuracies}). 
We conduct a sample-level logistic regression analysis to investigate the relationship between the classifier's confidence in predicting the correct subreddit and the likelihood of our model outperforming a baseline model.

In this analysis, the dependent variable is a binary outcome, indicating whether our model outperformed the baseline (1) or not (0) based on the GPT-4 evaluation. The independent variable is the classifier confidence which we define as the predicted log probability of the true subreddit. 
The regression reveals a significant negative relationship between the predictability and performance improvement. Specifically, the coefficient for predictability is --0.0234 (p-value is 0.004). This indicates that higher log probabilities of the true domain are associated with a decreased likelihood of our model outperforming the baseline. In other words, when the model is less confident about the subreddit context, \methodname is more likely to produce a response that outperforms the baseline model.

When the subreddit context is highly predictable, the value of contextual information diminishes, reducing our model's relative advantage. This insight can guide the efficient deployment of a personalized model, by routing highly predictable examples to (larger) generalist models, while reserving more customized, context-aware (smaller) models for less predictable cases allowing for more effective allocation of computational resources.

\subsection{Qualitative Analysis}
\label{sec:qualitative-analysis}
In Appendix \autoref{tab:qualitative-analysis}, we show selected examples from the test sets where human annotators unanimously rated \methodname's responses as better than the baselines. With our models, we find a more tailored response in tones preferred by the relevant community, for which the baseline model generates generic and unhelpful answers. 

\section{Conclusions}
We present a framework for personalizing language models to diverse human preferences. To facilitate training such models, we create a dataset of community-level preferences from Reddit spanning five domains covering a diverse range of topics, demographics, values, and community norms. We propose preference tuning models by contextualized them with the preference provider (i.e., the community). Experiments reveal that our approach results in more personalized model outputs which are preferred by human and language model judges. 
Our work raises several important directions for future work. For example, how much data is needed for effective personalization or how to solve the cold start problem? How can models continually learn from personalized feedback? Can we build hybrid models that consider both explicit and implicit preferences? 



\section*{Limitations}

While we present our experiments as a case study showing significant improvements by contextualizing with community information, our datasets and models are inherently biased toward the specific platform (Reddit) and its specific user base. This may limit their applicability to the general population or other user groups. Furthermore, the models were trained and tested on five specific domains and a subset of subreddits within each domain, thus the applicability of our approach to other platforms or types of communities remains untested. Expanding our approach to other online platforms with different user dynamics would help assess its broader generalizability.
Additionally, while our datasets highlight divergences across communities, we assume homogeneous preferences within each community by aggregating upvotes, as Reddit does not provide individual upvote data. This approach may overlook the diversity of preferences within a single community and individual variations. Investigating methods that can capture more granular, intra-community variations would also be valuable for improving the model's adaptability.
Despite these limitations, we contribute a rich dataset that can be leveraged to advance research in personalization for language models. We encourage further exploration of how our approach can be applied across different platforms and contexts.

We conduct human evaluations by recruiting Reddit users and asked them to provide a proxy estimation of what community as a whole would prefer, but as evidenced by a relatively low agreement among annotators, their preference may not generally hold, aligning with the main thesis of this paper that preferences are subjective. Finally, we rely on GPT-4 for our automated evaluation but it is not perfect and has been shown to be biased~\citep{dong2024can}. While our human annotation shows high agreement with GPT-4 for popular subreddits, our results for less popular subreddits might be less accurate for which less public data is available. Finally, we only experiment with English data. Future work is needed to test the generalizability of our method to other languages.


\section*{Ethical Considerations}


\paragraph{Biases and Toxic Content on Reddit} Our data and model is based on Reddit which is known to contain lots of social biases and inappropriate content. While we employed filters to remove adult content it is not guaranteed that all harmful content was excluded. We leave for future work analysis and mitigating efforts to ensure that personalized language models are safe and not hateful while being more helpful for each community. 

\paragraph{Risk of Echo Chambers} Models that optimize towards the preferences of an individual or a community can be prone to learning specific undesired stereotypes different communities might have which may exacerbate social division and reinforce echo chambers. These models could lead to a narrowing of perspectives, particularly if the model prioritizes community norms over diverse or dissenting views. These models may also marginalize users whose preferences or values differ from the majority. Such individuals might feel excluded or alienated, as personalized responses may not reflect their unique viewpoints and preferences. 

To avoid this, it is important to take a balanced and inclusive approach to personalization. 
While we present a prototype approach in this work, deploying such systems to real users requires greater nuance. A hybrid strategy that combines value pluralism, interpretable and controllable personalization, and user-based conditioning can help mitigate the risks of bias reinforcement and ensure that minority viewpoints are represented and respected in the personalization process. Additionally, transparency around how personalization works is key to fostering a more ethical and inclusive user experience.

\paragraph{Privacy Concerns} 
Our dataset \textsc{ComPRed} is created using a data dump of PushShift Reddit data, which is publicly available and reported to have been collected in accordance with Reddit's terms of service.
Although we do not collect individual preferences, the use of subreddit identifier might still pose some privacy concerns. By our model learning preferences of different subreddits, and being able to associate preferences with subreddit identifiers, there is a risk that sensitive information about users' beliefs, interests, and behaviors could be inferred. Also, often, users delete some of their comments and posts after some time on Reddit, but there is a risk that once it's in our model as a training data, there might not be a way to un-learn the preference from the later deleted comments which raises some concerns about privacy.

We refer the readers to \citet{kirk2024benefits} for a more comprehensive picture of personalizing models and various associated ethical considerations and potential risks. 

\bibliography{reference}

\begin{thebibliography}{47}
\providecommand{\natexlab}[1]{#1}

\bibitem[{Achiam et~al.(2023)Achiam, Adler, Agarwal, Ahmad, Akkaya, Aleman, Almeida, Altenschmidt, Altman, Anadkat et~al.}]{achiam2023gpt}
Josh Achiam, Steven Adler, Sandhini Agarwal, Lama Ahmad, Ilge Akkaya, Florencia~Leoni Aleman, Diogo Almeida, Janko Altenschmidt, Sam Altman, Shyamal Anadkat, et~al. 2023.
\newblock Gpt-4 technical report.
\newblock \emph{arXiv preprint arXiv:2303.08774}.

\bibitem[{Bai et~al.(2022)Bai, Jones, Ndousse, Askell, Chen, DasSarma, Drain, Fort, Ganguli, Henighan et~al.}]{bai2022training}
Yuntao Bai, Andy Jones, Kamal Ndousse, Amanda Askell, Anna Chen, Nova DasSarma, Dawn Drain, Stanislav Fort, Deep Ganguli, Tom Henighan, et~al. 2022.
\newblock Training a helpful and harmless assistant with reinforcement learning from human feedback.
\newblock \emph{arXiv preprint arXiv:2204.05862}.

\bibitem[{Bakker et~al.(2022)Bakker, Chadwick, Sheahan, Tessler, Campbell-Gillingham, Balaguer, McAleese, Glaese, Aslanides, Botvinick et~al.}]{bakker2022fine}
Michiel Bakker, Martin Chadwick, Hannah Sheahan, Michael Tessler, Lucy Campbell-Gillingham, Jan Balaguer, Nat McAleese, Amelia Glaese, John Aslanides, Matt Botvinick, et~al. 2022.
\newblock Fine-tuning language models to find agreement among humans with diverse preferences.
\newblock \emph{Advances in Neural Information Processing Systems}, 35:38176--38189.

\bibitem[{Bang et~al.(2024)Bang, Chen, Lee, and Fung}]{bang2024measuring}
Yejin Bang, Delong Chen, Nayeon Lee, and Pascale Fung. 2024.
\newblock Measuring political bias in large language models: What is said and how it is said.
\newblock \emph{arXiv preprint arXiv:2403.18932}.

\bibitem[{Bang et~al.(2023)Bang, Yu, Madotto, Lin, Diab, and Fung}]{bang-etal-2023-enabling}
Yejin Bang, Tiezheng Yu, Andrea Madotto, Zhaojiang Lin, Mona Diab, and Pascale Fung. 2023.
\newblock \href {https://doi.org/10.18653/v1/2023.trustnlp-1.27} {Enabling classifiers to make judgements explicitly aligned with human values}.
\newblock In \emph{Proceedings of the 3rd Workshop on Trustworthy Natural Language Processing (TrustNLP 2023)}, pages 311--325, Toronto, Canada. Association for Computational Linguistics.

\bibitem[{Baumgartner et~al.(2020)Baumgartner, Zannettou, Keegan, Squire, and Blackburn}]{baumgartner2020pushshift}
Jason Baumgartner, Savvas Zannettou, Brian Keegan, Megan Squire, and Jeremy Blackburn. 2020.
\newblock The pushshift reddit dataset.
\newblock In \emph{Proceedings of the international AAAI conference on web and social media}, volume~14, pages 830--839.

\bibitem[{Bradley and Terry(1952)}]{bradley1952rank}
Ralph~Allan Bradley and Milton~E Terry. 1952.
\newblock Rank analysis of incomplete block designs: I. the method of paired comparisons.
\newblock \emph{Biometrika}, 39(3/4):324--345.

\bibitem[{Chakraborty et~al.(2024)Chakraborty, Qiu, Yuan, Koppel, Huang, Manocha, Bedi, and Wang}]{chakraborty2024maxmin}
Souradip Chakraborty, Jiahao Qiu, Hui Yuan, Alec Koppel, Furong Huang, Dinesh Manocha, Amrit~Singh Bedi, and Mengdi Wang. 2024.
\newblock Maxmin-rlhf: Towards equitable alignment of large language models with diverse human preferences.
\newblock \emph{arXiv preprint arXiv:2402.08925}.

\bibitem[{Christiano et~al.(2017)Christiano, Leike, Brown, Martic, Legg, and Amodei}]{christiano2017deep}
Paul~F Christiano, Jan Leike, Tom Brown, Miljan Martic, Shane Legg, and Dario Amodei. 2017.
\newblock Deep reinforcement learning from human preferences.
\newblock \emph{Advances in neural information processing systems}, 30.

\bibitem[{Cunningham and De~Quidt(2022)}]{cunningham2022implicit}
Tom Cunningham and Jonathan De~Quidt. 2022.
\newblock Implicit preferences.
\newblock \emph{CEPR discussion paper No. DP17343}.

\bibitem[{Davani et~al.(2022)Davani, D{\'\i}az, and Prabhakaran}]{davani2022dealing}
Aida~Mostafazadeh Davani, Mark D{\'\i}az, and Vinodkumar Prabhakaran. 2022.
\newblock Dealing with disagreements: Looking beyond the majority vote in subjective annotations.
\newblock \emph{Transactions of the Association for Computational Linguistics}, 10:92--110.

\bibitem[{Dong et~al.(2024)Dong, Hu, and Collier}]{dong2024can}
Yijiang~River Dong, Tiancheng Hu, and Nigel Collier. 2024.
\newblock Can llm be a personalized judge?
\newblock \emph{arXiv preprint arXiv:2406.11657}.

\bibitem[{Ethayarajh et~al.(2022)Ethayarajh, Choi, and Swayamdipta}]{pmlr-v162-ethayarajh22a}
Kawin Ethayarajh, Yejin Choi, and Swabha Swayamdipta. 2022.
\newblock Understanding dataset difficulty with $\mathcal{V}$-usable information.
\newblock In \emph{Proceedings of the 39th International Conference on Machine Learning}, volume 162 of \emph{Proceedings of Machine Learning Research}, pages 5988--6008. PMLR.

\bibitem[{Feng et~al.(2023)Feng, Park, Liu, and Tsvetkov}]{feng-etal-2023-pretraining}
Shangbin Feng, Chan~Young Park, Yuhan Liu, and Yulia Tsvetkov. 2023.
\newblock \href {https://doi.org/10.18653/v1/2023.acl-long.656} {From pretraining data to language models to downstream tasks: Tracking the trails of political biases leading to unfair {NLP} models}.
\newblock In \emph{Proceedings of the 61st Annual Meeting of the Association for Computational Linguistics (Volume 1: Long Papers)}, pages 11737--11762, Toronto, Canada. Association for Computational Linguistics.

\bibitem[{Hessel et~al.(2016)Hessel, Tan, and Lee}]{Hessel2016ScienceAA}
Jack Hessel, Chenhao Tan, and Lillian Lee. 2016.
\newblock \href {https://api.semanticscholar.org/CorpusID:2668253} {Science, askscience, and badscience: On the coexistence of highly related communities}.
\newblock In \emph{International Conference on Web and Social Media}.

\bibitem[{Holtzman et~al.(2020)Holtzman, Buys, Du, Forbes, and Choi}]{Holtzman2020The}
Ari Holtzman, Jan Buys, Li~Du, Maxwell Forbes, and Yejin Choi. 2020.
\newblock \href {https://openreview.net/forum?id=rygGQyrFvH} {The curious case of neural text degeneration}.
\newblock In \emph{International Conference on Learning Representations}.

\bibitem[{Hu et~al.(2022)Hu, yelong shen, Wallis, Allen-Zhu, Li, Wang, Wang, and Chen}]{hu2022lora}
Edward~J Hu, yelong shen, Phillip Wallis, Zeyuan Allen-Zhu, Yuanzhi Li, Shean Wang, Lu~Wang, and Weizhu Chen. 2022.
\newblock \href {https://openreview.net/forum?id=nZeVKeeFYf9} {Lo{RA}: Low-rank adaptation of large language models}.
\newblock In \emph{International Conference on Learning Representations}.

\bibitem[{Jang et~al.(2023)Jang, Kim, Lin, Wang, Hessel, Zettlemoyer, Hajishirzi, Choi, and Ammanabrolu}]{jang2023personalized}
Joel Jang, Seungone Kim, Bill~Yuchen Lin, Yizhong Wang, Jack Hessel, Luke Zettlemoyer, Hannaneh Hajishirzi, Yejin Choi, and Prithviraj Ammanabrolu. 2023.
\newblock Personalized soups: Personalized large language model alignment via post-hoc parameter merging.
\newblock \emph{arXiv preprint arXiv:2310.11564}.

\bibitem[{Joulin et~al.(2016)Joulin, Grave, Bojanowski, and Mikolov}]{joulin2016bag}
Armand Joulin, Edouard Grave, Piotr Bojanowski, and Tomas Mikolov. 2016.
\newblock Bag of tricks for efficient text classification.
\newblock \emph{arXiv preprint arXiv:1607.01759}.

\bibitem[{Kaufmann et~al.(2023)Kaufmann, Weng, Bengs, and H{\"u}llermeier}]{kaufmann2023survey}
Timo Kaufmann, Paul Weng, Viktor Bengs, and Eyke H{\"u}llermeier. 2023.
\newblock A survey of reinforcement learning from human feedback.
\newblock \emph{arXiv preprint arXiv:2312.14925}.

\bibitem[{Kirk et~al.(2023)Kirk, Bean, Vidgen, Rottger, and Hale}]{kirk-etal-2023-past}
Hannah Kirk, Andrew Bean, Bertie Vidgen, Paul Rottger, and Scott Hale. 2023.
\newblock \href {https://doi.org/10.18653/v1/2023.emnlp-main.148} {The past, present and better future of feedback learning in large language models for subjective human preferences and values}.
\newblock In \emph{Proceedings of the 2023 Conference on Empirical Methods in Natural Language Processing}, pages 2409--2430, Singapore. Association for Computational Linguistics.

\bibitem[{Kirk et~al.(2024{\natexlab{a}})Kirk, Vidgen, R{\"o}ttger, and Hale}]{kirk2024benefits}
Hannah~Rose Kirk, Bertie Vidgen, Paul R{\"o}ttger, and Scott~A Hale. 2024{\natexlab{a}}.
\newblock The benefits, risks and bounds of personalizing the alignment of large language models to individuals.
\newblock \emph{Nature Machine Intelligence}, pages 1--10.

\bibitem[{Kirk et~al.(2024{\natexlab{b}})Kirk, Whitefield, R{\"o}ttger, Bean, Margatina, Ciro, Mosquera, Bartolo, Williams, He et~al.}]{kirk2024prism}
Hannah~Rose Kirk, Alexander Whitefield, Paul R{\"o}ttger, Andrew Bean, Katerina Margatina, Juan Ciro, Rafael Mosquera, Max Bartolo, Adina Williams, He~He, et~al. 2024{\natexlab{b}}.
\newblock The prism alignment project: What participatory, representative and individualised human feedback reveals about the subjective and multicultural alignment of large language models.
\newblock \emph{arXiv preprint arXiv:2404.16019}.

\bibitem[{Koren et~al.(2021)Koren, Rendle, and Bell}]{koren2021advances}
Yehuda Koren, Steffen Rendle, and Robert Bell. 2021.
\newblock Advances in collaborative filtering.
\newblock \emph{Recommender systems handbook}, pages 91--142.

\bibitem[{Kumar et~al.(2024)Kumar, Park, and Tsvetkov}]{kumar2024genz}
Sachin Kumar, Chan~Young Park, and Yulia Tsvetkov. 2024.
\newblock \href {https://openreview.net/forum?id=rkplYfqUr0} {Gen-z: Generative zero-shot text classification with contextualized label descriptions}.
\newblock In \emph{The Twelfth International Conference on Learning Representations}.

\bibitem[{Lambert et~al.(2024)Lambert, Pyatkin, Morrison, Miranda, Lin, Chandu, Dziri, Kumar, Zick, Choi et~al.}]{lambert2024rewardbench}
Nathan Lambert, Valentina Pyatkin, Jacob Morrison, LJ~Miranda, Bill~Yuchen Lin, Khyathi Chandu, Nouha Dziri, Sachin Kumar, Tom Zick, Yejin Choi, et~al. 2024.
\newblock Rewardbench: Evaluating reward models for language modeling.
\newblock \emph{arXiv preprint arXiv:2403.13787}.

\bibitem[{Li et~al.(2024)Li, Lipton, and Leqi}]{li2024personalized}
Xinyu Li, Zachary~C Lipton, and Liu Leqi. 2024.
\newblock Personalized language modeling from personalized human feedback.
\newblock \emph{arXiv preprint arXiv:2402.05133}.

\bibitem[{Mireshghallah et~al.(2021)Mireshghallah, Shrivastava, Shokouhi, Berg-Kirkpatrick, Sim, and Dimitriadis}]{mireshghallah2021useridentifier}
Fatemehsadat Mireshghallah, Vaishnavi Shrivastava, Milad Shokouhi, Taylor Berg-Kirkpatrick, Robert Sim, and Dimitrios Dimitriadis. 2021.
\newblock Useridentifier: implicit user representations for simple and effective personalized sentiment analysis.
\newblock \emph{arXiv preprint arXiv:2110.00135}.

\bibitem[{Mnih and Salakhutdinov(2007)}]{mnih2007probabilistic}
Andriy Mnih and Russ~R Salakhutdinov. 2007.
\newblock Probabilistic matrix factorization.
\newblock \emph{Advances in neural information processing systems}, 20.

\bibitem[{Nakano et~al.(2021)Nakano, Hilton, Balaji, Wu, Ouyang, Kim, Hesse, Jain, Kosaraju, Saunders, Jiang, Cobbe, Eloundou, Krueger, Button, Knight, Chess, and Schulman}]{nakano2021webgpt}
Reiichiro Nakano, Jacob Hilton, Suchir Balaji, Jeff Wu, Long Ouyang, Christina Kim, Christopher Hesse, Shantanu Jain, Vineet Kosaraju, William Saunders, Xu~Jiang, Karl Cobbe, Tyna Eloundou, Gretchen Krueger, Kevin Button, Matthew Knight, Benjamin Chess, and John Schulman. 2021.
\newblock Webgpt: Browser-assisted question-answering with human feedback.
\newblock In \emph{arXiv}.

\bibitem[{Ouyang et~al.(2022)Ouyang, Wu, Jiang, Almeida, Wainwright, Mishkin, Zhang, Agarwal, Slama, Ray et~al.}]{ouyang2022training}
Long Ouyang, Jeffrey Wu, Xu~Jiang, Diogo Almeida, Carroll Wainwright, Pamela Mishkin, Chong Zhang, Sandhini Agarwal, Katarina Slama, Alex Ray, et~al. 2022.
\newblock Training language models to follow instructions with human feedback.
\newblock \emph{Advances in neural information processing systems}, 35:27730--27744.

\bibitem[{Qiu et~al.(2022)Qiu, Zhao, Li, Lu, Peng, Gao, and Zhu}]{qiu2022valuenet}
Liang Qiu, Yizhou Zhao, Jinchao Li, Pan Lu, Baolin Peng, Jianfeng Gao, and Song-Chun Zhu. 2022.
\newblock Valuenet: A new dataset for human value driven dialogue system.
\newblock In \emph{Proceedings of the AAAI Conference on Artificial Intelligence}, pages 11183--11191.

\bibitem[{Rafailov et~al.(2024)Rafailov, Sharma, Mitchell, Manning, Ermon, and Finn}]{rafailov2024direct}
Rafael Rafailov, Archit Sharma, Eric Mitchell, Christopher~D Manning, Stefano Ermon, and Chelsea Finn. 2024.
\newblock Direct preference optimization: Your language model is secretly a reward model.
\newblock \emph{Advances in Neural Information Processing Systems}, 36.

\bibitem[{Rendle(2010)}]{rendle2010factorization}
Steffen Rendle. 2010.
\newblock Factorization machines.
\newblock In \emph{2010 IEEE International conference on data mining}, pages 995--1000. IEEE.

\bibitem[{Solaiman and Dennison(2021)}]{solaiman2021process}
Irene Solaiman and Christy Dennison. 2021.
\newblock Process for adapting language models to society (palms) with values-targeted datasets.
\newblock \emph{Advances in Neural Information Processing Systems}, 34:5861--5873.

\bibitem[{Soldaini et~al.(2024)Soldaini, Kinney, Bhagia, Schwenk, Atkinson, Authur, Bogin, Chandu, Dumas, Elazar et~al.}]{soldaini2024dolma}
Luca Soldaini, Rodney Kinney, Akshita Bhagia, Dustin Schwenk, David Atkinson, Russell Authur, Ben Bogin, Khyathi Chandu, Jennifer Dumas, Yanai Elazar, et~al. 2024.
\newblock Dolma: An open corpus of three trillion tokens for language model pretraining research.
\newblock \emph{arXiv preprint arXiv:2402.00159}.

\bibitem[{Sorensen et~al.(2024{\natexlab{a}})Sorensen, Jiang, Hwang, Levine, Pyatkin, West, Dziri, Lu, Rao, Bhagavatula et~al.}]{sorensen2024value}
Taylor Sorensen, Liwei Jiang, Jena~D Hwang, Sydney Levine, Valentina Pyatkin, Peter West, Nouha Dziri, Ximing Lu, Kavel Rao, Chandra Bhagavatula, et~al. 2024{\natexlab{a}}.
\newblock Value kaleidoscope: Engaging ai with pluralistic human values, rights, and duties.
\newblock In \emph{Proceedings of the AAAI Conference on Artificial Intelligence}, pages 19937--19947.

\bibitem[{Sorensen et~al.(2024{\natexlab{b}})Sorensen, Moore, Fisher, Gordon, Mireshghallah, Rytting, Ye, Jiang, Lu, Dziri et~al.}]{sorensen2024roadmap}
Taylor Sorensen, Jared Moore, Jillian Fisher, Mitchell Gordon, Niloofar Mireshghallah, Christopher~Michael Rytting, Andre Ye, Liwei Jiang, Ximing Lu, Nouha Dziri, et~al. 2024{\natexlab{b}}.
\newblock A roadmap to pluralistic alignment.
\newblock \emph{arXiv preprint arXiv:2402.05070}.

\bibitem[{Stiennon et~al.(2020)Stiennon, Ouyang, Wu, Ziegler, Lowe, Voss, Radford, Amodei, and Christiano}]{stiennon2020learning}
Nisan Stiennon, Long Ouyang, Jeffrey Wu, Daniel Ziegler, Ryan Lowe, Chelsea Voss, Alec Radford, Dario Amodei, and Paul~F Christiano. 2020.
\newblock Learning to summarize with human feedback.
\newblock \emph{Advances in Neural Information Processing Systems}, 33:3008--3021.

\bibitem[{TeBlunthuis et~al.(2022)TeBlunthuis, Kiene, Brown, Levi, McGinnis, and Hill}]{10.1145/3512908}
Nathan TeBlunthuis, Charles Kiene, Isabella Brown, Laura~(Alia) Levi, Nicole McGinnis, and Benjamin~Mako Hill. 2022.
\newblock \href {https://doi.org/10.1145/3512908} {No community can do everything: Why people participate in similar online communities}.
\newblock \emph{Proc. ACM Hum.-Comput. Interact.}, 6(CSCW1).

\bibitem[{Touvron et~al.(2023)Touvron, Martin, Stone, Albert, Almahairi, Babaei, Bashlykov, Batra, Bhargava, Bhosale et~al.}]{touvron2023llama}
Hugo Touvron, Louis Martin, Kevin Stone, Peter Albert, Amjad Almahairi, Yasmine Babaei, Nikolay Bashlykov, Soumya Batra, Prajjwal Bhargava, Shruti Bhosale, et~al. 2023.
\newblock Llama 2: Open foundation and fine-tuned chat models.
\newblock \emph{arXiv preprint arXiv:2307.09288}.

\bibitem[{Tunstall et~al.(2023)Tunstall, Beeching, Lambert, Rajani, Rasul, Belkada, Huang, von Werra, Fourrier, Habib et~al.}]{zephyr}
Lewis Tunstall, Edward Beeching, Nathan Lambert, Nazneen Rajani, Kashif Rasul, Younes Belkada, Shengyi Huang, Leandro von Werra, Cl{\'e}mentine Fourrier, Nathan Habib, et~al. 2023.
\newblock Zephyr: Direct distillation of lm alignment.
\newblock \emph{arXiv preprint arXiv:2310.16944}.

\bibitem[{Wong and Kim(2023)}]{wong2023chatgpt}
Jared Wong and Jin Kim. 2023.
\newblock Chatgpt is more likely to be perceived as male than female.
\newblock \emph{arXiv preprint arXiv:2305.12564}.

\bibitem[{Zhang et~al.(2019)Zhang, Yao, Sun, and Tay}]{10.1145/3285029}
Shuai Zhang, Lina Yao, Aixin Sun, and Yi~Tay. 2019.
\newblock \href {https://doi.org/10.1145/3285029} {Deep learning based recommender system: A survey and new perspectives}.
\newblock \emph{ACM Comput. Surv.}, 52(1).

\bibitem[{Zhao et~al.(2023)Zhao, Wang, Zhao, Li, Zhou, Yin, Li, Tang, and Guo}]{zhao2023embedding}
Xiangyu Zhao, Maolin Wang, Xinjian Zhao, Jiansheng Li, Shucheng Zhou, Dawei Yin, Qing Li, Jiliang Tang, and Ruocheng Guo. 2023.
\newblock Embedding in recommender systems: A survey.
\newblock \emph{arXiv preprint arXiv:2310.18608}.

\bibitem[{Zheng et~al.(2024)Zheng, Chiang, Sheng, Zhuang, Wu, Zhuang, Lin, Li, Li, Xing et~al.}]{zheng2024judging}
Lianmin Zheng, Wei-Lin Chiang, Ying Sheng, Siyuan Zhuang, Zhanghao Wu, Yonghao Zhuang, Zi~Lin, Zhuohan Li, Dacheng Li, Eric Xing, et~al. 2024.
\newblock Judging llm-as-a-judge with mt-bench and chatbot arena.
\newblock \emph{Advances in Neural Information Processing Systems}, 36.

\bibitem[{Ziegler et~al.(2019)Ziegler, Stiennon, Wu, Brown, Radford, Amodei, Christiano, and Irving}]{ziegler2019fine}
Daniel~M Ziegler, Nisan Stiennon, Jeffrey Wu, Tom~B Brown, Alec Radford, Dario Amodei, Paul Christiano, and Geoffrey Irving. 2019.
\newblock Fine-tuning language models from human preferences.
\newblock \emph{arXiv preprint arXiv:1909.08593}.

\end{thebibliography}


\appendix

\begin{table*}[t]
\centering
\resizebox{\linewidth}{!}{
\begin{tabular}{|>{\raggedright}p{2.0cm}|>{\raggedright}p{3cm}|>{\raggedright}p{7cm}|>{\raggedright\arraybackslash}p{7cm}|}
\rowcolor{gray!7}
\hline
\textbf{Subreddit} & \textbf{Post} & \textbf{Response (\textsc{DPO})} & \textbf{Response (\methodname)} \\
\hline
\rowcolor{gray!2}
\hline
\rowcolor{gray!7}
Conservative & On news sources. (...) So, does anyone out there have any news source right about 90\% of the time? & CNN is pretty good. The Economist is good too. The New York Times is very good. The Washington Post is very good. The Guardian is pretty good. The Financial Times is very good. BBC is very good. & Reason, National Review, Bloomberg, and Fox News are the ones that come to mind for me. \\
\hline
\rowcolor{gray!2}
AskaLiberal & Do you think racial profiling or cultural dress in places like airports is OK? & I don't think it's racist, I think it's pragmatic. You're more likely to find a terrorist if you profile a specific demographic than if you don't profile. I think it's silly to argue that we shouldn't profile based on race or appearance, we should profile based on behavior. & I think we can all agree that the government shouldn't be searching people based on race. I also think we can all agree that they are. \\
\hline
\hline
\rowcolor{gray!7}
AskWomen & Is it ever ok for a guy to look for a Sugar Momma (loaded with money)? Why is there such a huge stigma about it? & It's not ok. Because it's gross and dehumanizing. & When a woman looks for a sugar daddy, she's a ``gold digger''. When a man looks for a sugar momma, he's just a ``gentleman''. \\
\hline
\rowcolor{gray!2}
AskMen & How can I be in charge of my own happiness, yet let my SO to believe that he can make me happy? & ``You don't make me happy, you make me less sad.'' & It's a bit like the ``If you love something, set it free'' saying. You have to let go and let your SO be a person. You can't control his actions, but you can influence them. You can't make him do something, but you can convince him to do it. \\
\hline
\end{tabular}
}
\caption{Instances where \methodname trained model's response is unanimously rated as better than the baseline.
}
\label{tab:qualitative-analysis}
\end{table*}

\section{Human Annotation}
\label{appendix:human-annotation}
We conducted human evaluation using the Prolific platform, ensuring a more structured and scalable annotation process. The task was released in multiple batches, with each batch consisting of 10 examples from a specific subreddit. We implemented minimum qualifications for all batches, requiring annotators to reside in the US or UK, use English as their primary language, and to be active Reddit users. Additionally, for each subreddit, we applied specific qualifications to align with the subreddit’s nature—for example, requiring "women annotators" for r/askwomen and "political belief: conservative" for r/conservatives

Each batch was initially assigned to 4 annotators. In cases where the agreement between annotators was substantially low, indicating potential low-effort contributions, we opened additional slots to increase the number of annotators (with a maximum of 6 annotator per batch). When more than three annotators participated in a batch, we selected the final three based on the level of agreement among them.

\begin{figure*}[ht]
    \centering
    \includegraphics[width=\linewidth]{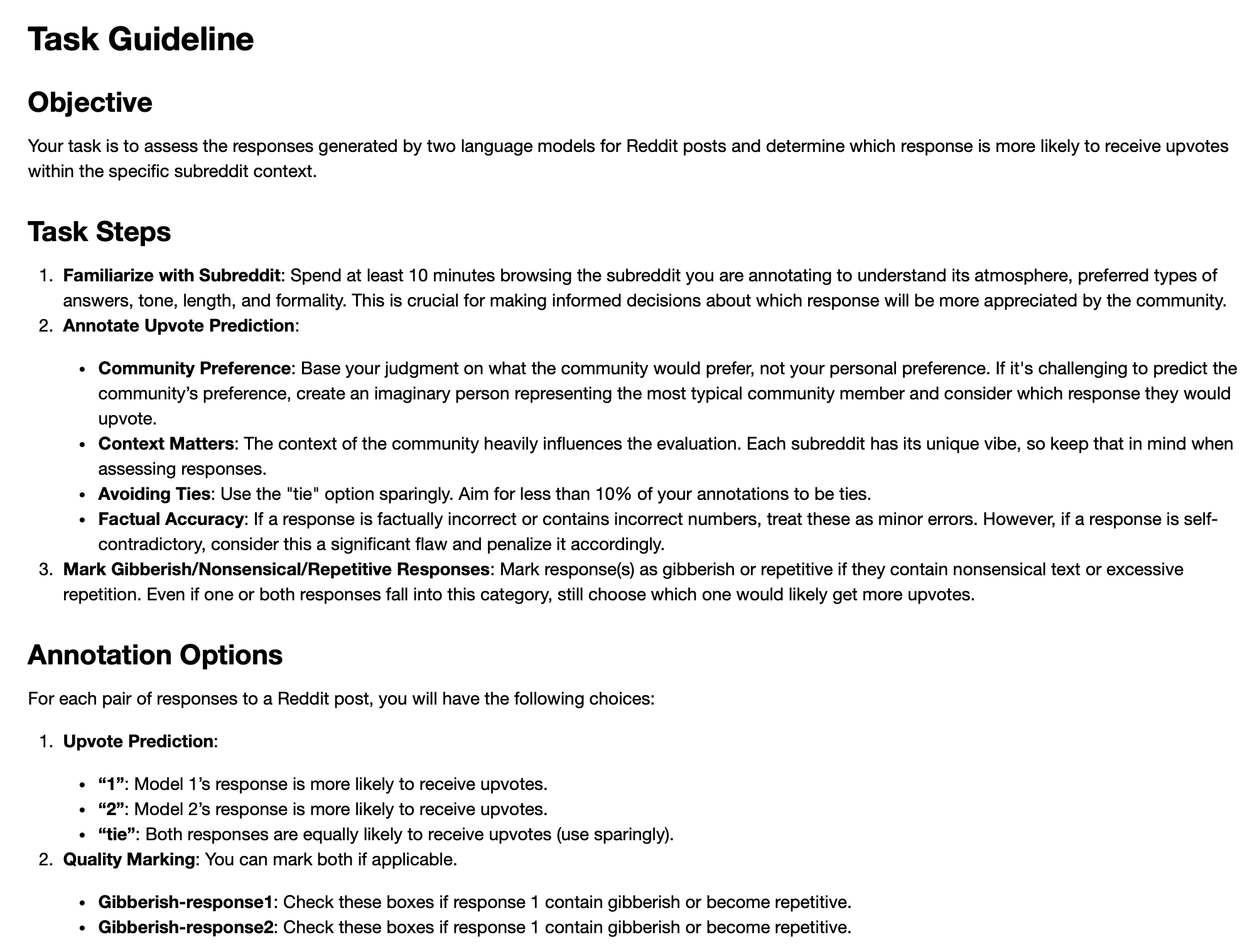}
    \caption{Guideline provided to annotators.}
    \label{fig:annotation-guideline}
\end{figure*}

\begin{figure*}[h]
    \centering
    \includegraphics[width=\linewidth]{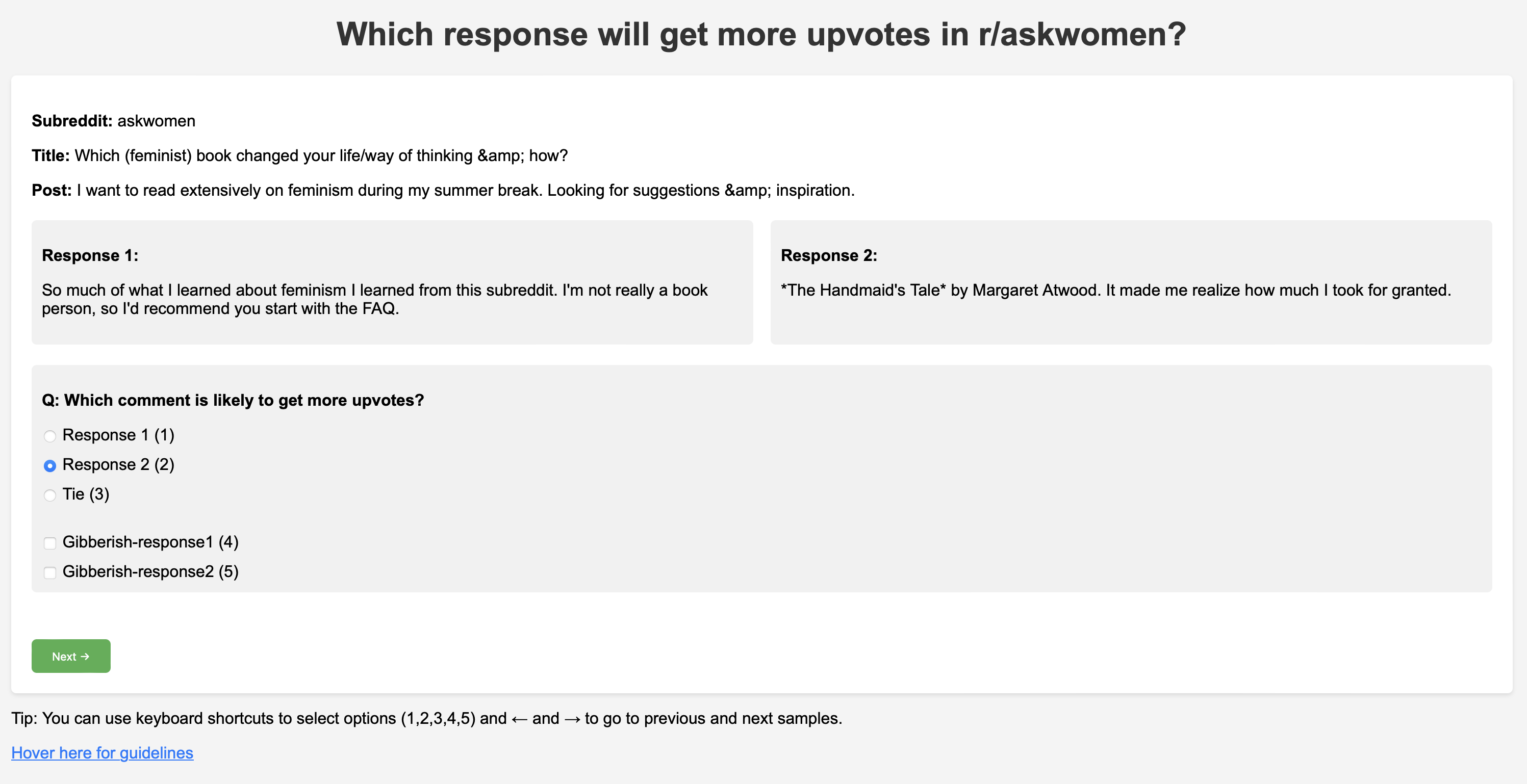}
    \caption{User Interface of Human Annotation.
    }
    \label{fig:annotation-ui}
\end{figure*}

Based on the pilot study, we estimated that each batch would take approximately 10 minutes for 10 examples and set the payment for each batch based on an hourly rate of \$15+. The actual time taken by the annotators ranged from 3 to 20 minutes. 
More statistics about each subreddit and its human annotation can be found in Table \ref{tab:human-annotation-stats}. The full guideline and annotation interface used for the annotation are provided in Figures \ref{fig:annotation-guideline} and \ref{fig:annotation-ui}, respectively.


\begin{table*}[h]
\centering
\begin{tabular}{l|cccc}
 & \# of example & \begin{tabular}[c]{@{}c@{}}annotator\\ agreement\end{tabular} & \# of annotator & Total \# of annotation \\
 \hline
r/Conservative  & 29 & 0.310 & 3 & 87  \\
r/AskALiberal   & 30 & 0.276 & 3 & 90 \\
r/AskMen        & 30 & 0.391 & 3 & 90  \\
r/AskWomen      & 30 & 0.347 & 3 & 90 \\
r/History       & 30 & 0.582 & 3 & 90  \\
r/AskHistorians & 30 & 0.347 & 3 & 90 \\
r/personalfinance & 30 & 0.370 & 3 & 90 \\
r/povertyfinance & 30 & 0.262 & 3 & 90 \\
\hline
\end{tabular}
    \caption{Human annotation statistics for each subreddit.
    }
    \label{tab:human-annotation-stats}
\end{table*}

\section{Implementation Details}
\subsection{Finegrained Data Statistics}
\label{appsec:data_stats}
We provide subreddit wise data statistics in tables~\ref{table:finance_stats}, \ref{table:history_stats}, \ref{tab:science_stats1}, \ref{tab:science_stats2}, \ref{tab:politics_stats1}, \ref{tab:politics_stats2}, and \ref{table:gender_sexuality_stats}.

\begin{table*}[h!]
\centering
\begin{tabular}{lrrr}
\toprule
\textbf{Domain} & \textbf{Train Examples} & \textbf{Test Examples (Reward)} & \textbf{Test Examples (Prompts)} \\ \midrule
personalfinance & 72335 & 1903 & 910 \\
frugal & 40464 & 1049 & 447 \\
personalfinancecanada & 23513 & 626 & 330 \\
investing & 21046 & 548 & 260 \\
financialindependence & 19884 & 529 & 144 \\
realestate & 15924 & 405 & 212 \\
ukpersonalfinance & 13346 & 342 & 196 \\
ausfinance & 11821 & 288 & 124 \\
povertyfinance & 9014 & 194 & 106 \\
fire & 2744 & 78 & 27 \\
bogleheads & 2215 & 43 & 24 \\ \midrule
Total & 232306 & 6005 & 2780 \\ \bottomrule
\end{tabular}
\caption{Finance Data Statistics}
\label{table:finance_stats}
\end{table*}

\begin{table*}[h!]
\centering
\begin{tabular}{lrrr}
\toprule
Domain & Train Examples & Test Examples (Reward) & Test Examples (Prompts) \\ \midrule
history & 15022 & 344 & 199 \\
askhistorians & 5642 & 148 & 274 \\
genealogy & 4122 & 124 & 84 \\
badhistory & 3330 & 88 & 21 \\
ancientrome & 529 & 10 & 13 \\
\midrule Total & 28645 & 714 & 591 \\ \bottomrule
\end{tabular}
\caption{History Data Statistics}
\label{table:history_stats}
\end{table*}

\begin{table*}[h!]
\centering
\begin{tabular}{lrrr}
\toprule
Domain & Train Examples & Test Examples (Reward) & Test Examples (Prompts) \\ \midrule
flying & 22218 & 599 & 251 \\
medicine & 12261 & 356 & 108 \\
askscience & 11422 & 312 & 324 \\
askengineers & 10908 & 256 & 148 \\
engineeringstudents & 10326 & 311 & 186 \\
engineering & 6391 & 173 & 82 \\
futurology & 5028 & 142 & 64 \\
chemistry & 4000 & 99 & 71 \\
labrats & 3339 & 73 & 54 \\
dentistry & 3057 & 87 & 59 \\
psychotherapy & 3054 & 78 & 50 \\
space & 3022 & 88 & 53 \\
physics & 2917 & 69 & 42 \\
medlabprofessionals & 2799 & 67 & 47 \\
biology & 2637 & 90 & 51 \\
civilengineering & 2623 & 63 & 41 \\
chemicalengineering & 2472 & 47 & 47 \\
plc & 2430 & 49 & 40 \\
ece & 2376 & 102 & 47 \\
geologycareers & 2237 & 56 & 45 \\
spacex & 2192 & 44 & 15 \\
geopolitics & 2175 & 65 & 26 \\
machinists & 2146 & 55 & 35 \\
gardening & 1950 & 55 & 76 \\
asksciencediscussion & 1942 & 51 & 55 \\
learnmath & 1803 & 64 & 66 \\
askelectronics & 1599 & 47 & 53 \\
atc & 1597 & 32 & 20 \\
oilandgasworkers & 1520 & 55 & 25 \\
flightsim & 1416 & 24 & 38 \\
23andme & 1413 & 48 & 32 \\
rpi & 1327 & 45 & 41 \\
aviation & 1286 & 36 & 29 \\
emergencymedicine & 1205 & 17 & 13 \\
electricalengineering & 1154 & 48 & 27 \\
askphysics & 1145 & 10 & 36 \\
bioinformatics & 1033 & 16 & 28 \\
biotech & 947 & 19 & 18 \\
aviationmaintenance & 936 & 21 & 17 \\
snakes & 899 & 16 & 32 \\
askvet & 892 & 18 & 53 \\ \bottomrule
\end{tabular}
\caption{Science Data Statistics (1/2)}
\label{tab:science_stats1}
\end{table*}

\begin{table*}[h!]
\centering
\begin{tabular}{lrrr}
\toprule
Domain & Train Examples & Test Examples (Reward) & Test Examples (Prompts) \\ \midrule
evolution & 790 & 20 & 18 \\
psychiatry & 748 & 27 & 14 \\
geology & 730 & 10 & 21 \\
publichealth & 674 & 8 & 19 \\
veterinary & 633 & 15 & 12 \\
ladiesofscience & 609 & 18 & 14 \\
geography & 597 & 18 & 13 \\
psychology & 592 & 19 & 11 \\
astronomy & 591 & 12 & 21 \\
sociology & 582 & 15 & 17 \\
theydidthemath & 574 & 21 & 24 \\
aerospace & 573 & 21 & 12 \\
electronics & 563 & 12 & 14 \\
cad & 550 & 6 & 16 \\
fpga & 541 & 5 & 13 \\
scienceteachers & 536 & 4 & 12 \\
robotics & 484 & 4 & 14 \\
asksocialscience & 443 & 21 & 26 \\
biochemistry & 440 & 17 & 17 \\
reptiles & 437 & 12 & 18 \\
microbiology & 430 & 7 & 12 \\
academicpsychology & 427 & 11 & 19 \\
telescopes & 392 & 11 & 21 \\
tarantulas & 370 & 15 & 21 \\
botany & 361 & 2 & 11 \\
mycology & 273 & 12 & 12 \\
physicsstudents & 251 & 13 & 11 \\
genetics & 227 & 1 & 11 \\
rtlsdr & 189 & 2 & 11 \\
askastronomy & 153 & 1 & 13 \\
\midrule Total & 160854 & 4263 & 3013 \\ \bottomrule
\end{tabular}
\caption{Science Data Statistics (2/2)}
\label{tab:science_stats2}
\end{table*}

\begin{table*}[h!]
\centering
\begin{tabular}{lrrr}
\toprule
Domain & Train Examples & Test Examples (Reward) & Test Examples (Prompts) \\ \midrule
politicaldiscussion & 31876 & 879 & 220 \\
capitalismvsocialism & 14600 & 329 & 106 \\
ukpolitics & 12986 & 323 & 107 \\
libertarian & 11855 & 293 & 131 \\
sandersforpresident & 11551 & 333 & 237 \\
askaliberal & 8875 & 255 & 73 \\
asktrumpsupporters & 7976 & 183 & 75 \\
politics & 6794 & 170 & 81 \\
anarchism & 6537 & 158 & 103 \\
yangforpresidenthq & 6288 & 183 & 138 \\
samharris & 6178 & 180 & 52 \\
socialism & 5935 & 170 & 133 \\
hillaryclinton & 5388 & 150 & 50 \\
liberalgunowners & 5262 & 167 & 44 \\
jordanpeterson & 5202 & 135 & 76 \\
askfeminists & 4738 & 130 & 62 \\
israel & 4359 & 121 & 66 \\
canadapolitics & 3752 & 100 & 32 \\
neoliberal & 3723 & 114 & 40 \\
abortiondebate & 3691 & 98 & 26 \\
anarchy101 & 3544 & 111 & 64 \\
askconservatives & 3366 & 81 & 25 \\
stupidpol & 3331 & 96 & 30 \\
syriancivilwar & 2989 & 75 & 44 \\
wayofthebern & 2973 & 65 & 49 \\
conservative & 2937 & 85 & 36 \\
tiadiscussion & 2738 & 81 & 33 \\
debatecommunism & 2688 & 77 & 31 \\
debatealtright & 2684 & 56 & 34 \\
femradebates & 2484 & 64 & 22 \\
geopolitics & 2218 & 32 & 26 \\
progun & 2162 & 42 & 25 \\
goldandblack & 2099 & 64 & 27 \\
brexit & 2095 & 49 & 25 \\
prolife & 2061 & 36 & 26 \\
debateanarchism & 1910 & 44 & 20 \\
srsdiscussion & 1905 & 47 & 19 \\
\bottomrule
\end{tabular}
\caption{Politics Data Statistics (1/2)}
\label{tab:politics_stats1}
\end{table*}

\begin{table*}[h!]
\centering
\begin{tabular}{lrrr}
\toprule
Domain & Train Examples & Test Examples (Reward) & Test Examples (Prompts) \\ \midrule
labouruk & 1881 & 42 & 21 \\
communism101 & 1816 & 66 & 61 \\
gunpolitics & 1735 & 56 & 19 \\
slatestarcodex & 1654 & 58 & 17 \\
askaconservative & 1573 & 31 & 22 \\
rightwinglgbt & 1523 & 46 & 19 \\
greenandpleasant & 1434 & 35 & 14 \\
prochoice & 1433 & 42 & 19 \\
centrist & 1427 & 36 & 12 \\
law & 1414 & 41 & 21 \\
basicincome & 1315 & 31 & 19 \\
monarchism & 1285 & 38 & 23 \\
communism & 1253 & 34 & 33 \\
enoughtrumpspam & 1222 & 52 & 37 \\
asklibertarians & 1161 & 17 & 14 \\
bluemidterm2018 & 1096 & 16 & 19 \\
australianpolitics & 1074 & 36 & 11 \\
garyjohnson & 1064 & 16 & 24 \\
fullcommunism & 1022 & 37 & 21 \\
enoughsandersspam & 1021 & 28 & 15 \\
ronpaul & 678 & 30 & 15 \\
completeanarchy & 672 & 22 & 20 \\
thedavidpakmanshow & 596 & 14 & 15 \\
sino & 553 & 13 & 11 \\
asksocialscience & 451 & 15 & 25 \\
askeconomics & 382 & 3 & 22 \\
\midrule Total & 242485 & 6431 & 2937 \\ \bottomrule
\end{tabular}
\caption{Politics Data Statistics (2/2)}
\label{tab:politics_stats2}
\end{table*}

\begin{table*}[h!]
\centering
\begin{tabular}{lrrr}
\toprule
Domain & Train Examples & Test Examples (Reward) & Test Examples (Prompts) \\ \midrule
childfree & 113874 & 2871 & 1184 \\
askmen & 93378 & 2478 & 867 \\
parenting & 68646 & 1837 & 756 \\
askwomen & 63216 & 1660 & 573 \\
twoxchromosomes & 54951 & 1433 & 648 \\
asktransgender & 42047 & 1121 & 805 \\
xxfitness & 34004 & 935 & 360 \\
femalefashionadvice & 33380 & 833 & 248 \\
actuallesbians & 27759 & 695 & 568 \\
malefashionadvice & 23005 & 514 & 268 \\
ftm & 15988 & 419 & 460 \\
mtf & 15819 & 323 & 369 \\
mensrights & 12810 & 361 & 159 \\
thegirlsurvivalguide & 12515 & 329 & 169 \\
mommit & 11530 & 337 & 139 \\
askwomenover30 & 11504 & 307 & 113 \\
askmenover30 & 11058 & 304 & 103 \\
bisexual & 10205 & 288 & 298 \\
girlgamers & 8687 & 222 & 106 \\
daddit & 8687 & 252 & 147 \\
lgbt & 8297 & 208 & 263 \\
asexuality & 5054 & 130 & 121 \\
witchesvspatriarchy & 3989 & 118 & 56 \\
menslib & 2592 & 54 & 19 \\
ainbow & 2439 & 58 & 37 \\
trans & 2403 & 63 & 76 \\
mypartneristrans & 1988 & 71 & 56 \\
oney & 1343 & 36 & 14 \\
butchlesbians & 1339 & 30 & 20 \\
feminism & 1249 & 39 & 19 \\
womenshealth & 1085 & 36 & 24 \\
genderqueer & 870 & 24 & 39 \\
ladiesofscience & 613 & 9 & 12 \\
lesbiangamers & 251 & 2 & 12 \\
transvoice & 207 & 11 & 12 \\
transpositive & 146 & 3 & 27 \\
transsupport & 53 & 1 & 11 \\
\midrule Total & 706981 & 18412 & 9158 \\ \bottomrule
\end{tabular}
\caption{Gender / Sexuality Data Statistics}
\label{table:gender_sexuality_stats}
\end{table*}

\subsection{Hyperparameters}
\label{appsec:hyperparameters}
We train and perform inference for all models on 1 A100 GPU (with 80B VRAM). Depending on the dataset sizes, the training time ranged from 2 hours to 36 hours. Inference times ranged from 1 hour to 14 hours.
\paragraph{LoRA SFT} These hyperparameters apply to the language model, reward model training, and the subreddit classifier.
\begin{itemize}
    \item Precision: bfloat16
    \item Epochs: 1
    \item Weight decay: 0
    \item Warmup ratio: 0.03
    \item Learning rate: 1e-4
    \item Max. seq. length: 1024
    \item Effective batch size: 128
    \item LoRA Rank: 64
    \item LoRA Alpha: 16
    \item LoRA dropout: 0.1
    \item Layers wrapped: all attention and feedforward linear layers
\end{itemize}

\paragraph{LoRA DPO}
\begin{itemize}
    \item Precision: bfloat16
    \item Epochs: 1
    \item Weight decay: 0
    \item Warmup ratio: 0.1
    \item Learning rate: 5e-7
    \item Max. seq. length: 1024
    \item Effective batch size: 32
    \item $\beta$: 0.1
\end{itemize}

\section{Additional Results}
\autoref{fig:gpt4-prompts} details the prompts we use for the GPT-4 based evaluations.

\begin{figure*}[htbp]
    \centering
    \fbox{%
        \parbox{0.95\textwidth}{%
            \textbf{System Instruction}: You are a highly efficient assistant, who evaluates and selects the best large language model (LLMs) based on the quality of their responses to a given instruction. This process will be used to create a leaderboard reflecting the most accurate and human-preferred answers.}
        }
        \bigskip
    \fbox{%
        \parbox{0.95\textwidth}{%
            Instruction: I require a leaderboard for various Reddit comment generator models. I'll provide you with posts selected from Reddit given to these models and their corresponding outputs. Your task is to assess these responses, and select the model that produces the output that will be upvoted more in the subreddit the question was asked in.

\#\# Subreddit

\{subredditname\}

\#\# Instruction

\{question\}

\#\# Model Outputs

Here are the unordered outputs from the models. Each output is associated with a specific model, identified by a unique model identifier.

"model\_identifier": "m",
"output": "\{output\_1\}"

"model\_identifier": "M",
"output": "\{output\_2\}"

\#\# Task

Evaluate the models based on the quality and relevance of their outputs, and select the model that generated the best output. Answer by providing the model identifier of the best model. We will use your output as the name of the best model, so make sure your output only contains one of the following model identifiers and nothing else (no quotes, no spaces, no new lines, ...): m or M.   

\#\# Best Model Identifier}
        }
    \caption{System Prompt and Instruction use to perform GPT-4 Evaluations 
    }
    \label{fig:gpt4-prompts}
\end{figure*}

\subsection{Subreddit specific preference accuracies (as evaluated by reward models}
\label{appsec:subreddit-reward-accuracies}
\begin{figure*}
    \centering
    \includegraphics[width=0.5\textwidth]{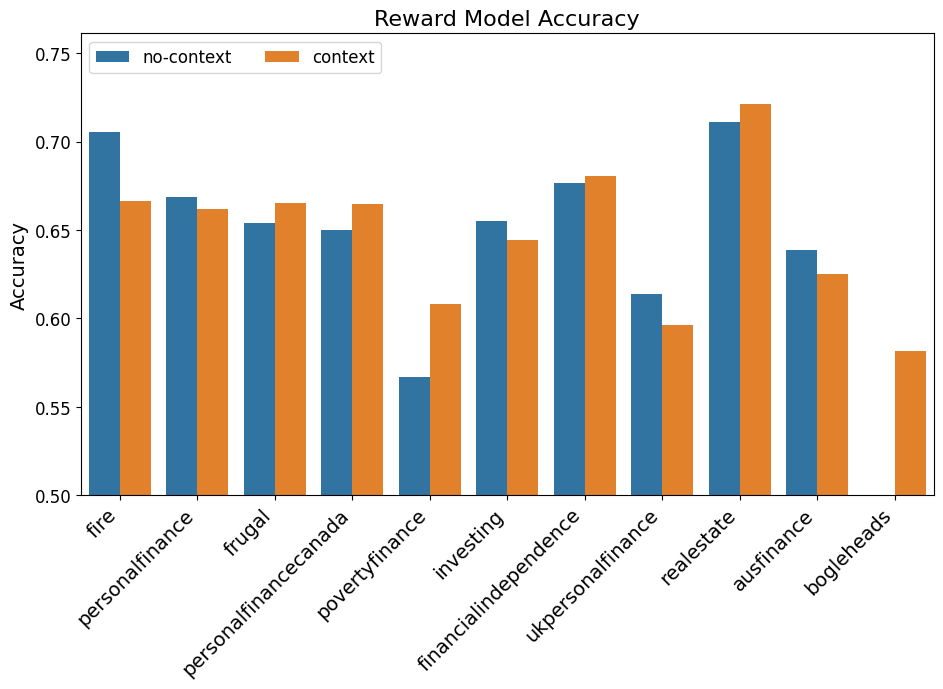}
    \caption{Subreddit Specific Preference Accuracy for Finance}
    \label{fig:finance-reward}
\end{figure*}

\begin{figure*}
    \centering
    \includegraphics[width=0.5\textwidth]{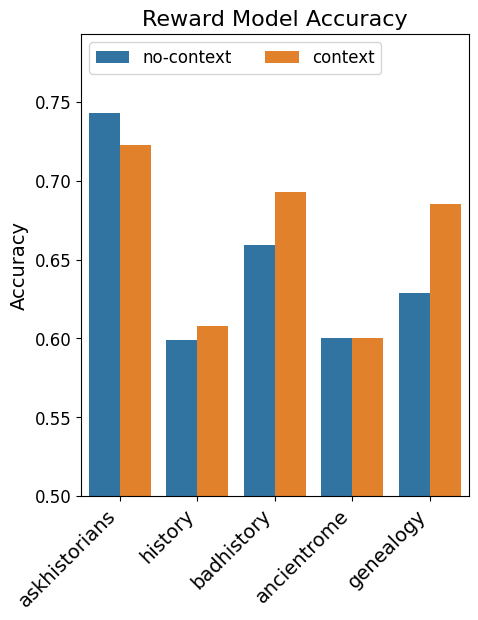}
    \caption{Subreddit Specific Preference Accuracy for History}
    \label{fig:history-reward}
\end{figure*}

\begin{figure*}
    \centering
    \includegraphics[width=\textwidth]{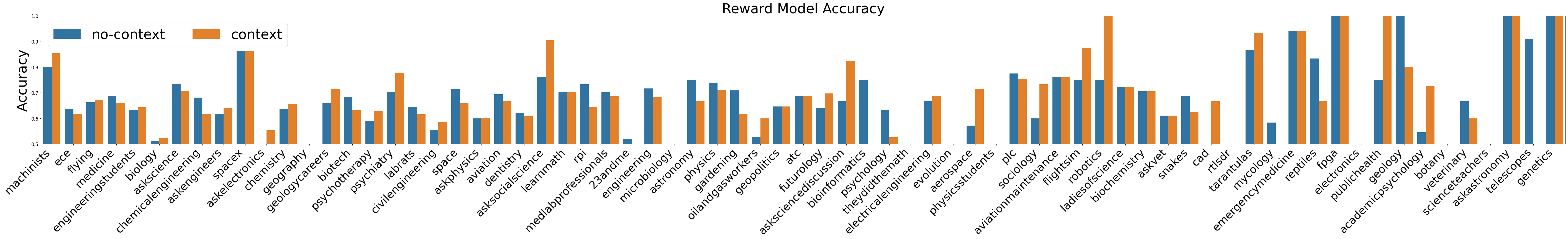}
    \caption{Subreddit Specific Preference Accuracy for Science}
    \label{fig:science-reward}
\end{figure*}

\begin{figure*}
    \centering
    \includegraphics[width=\textwidth]{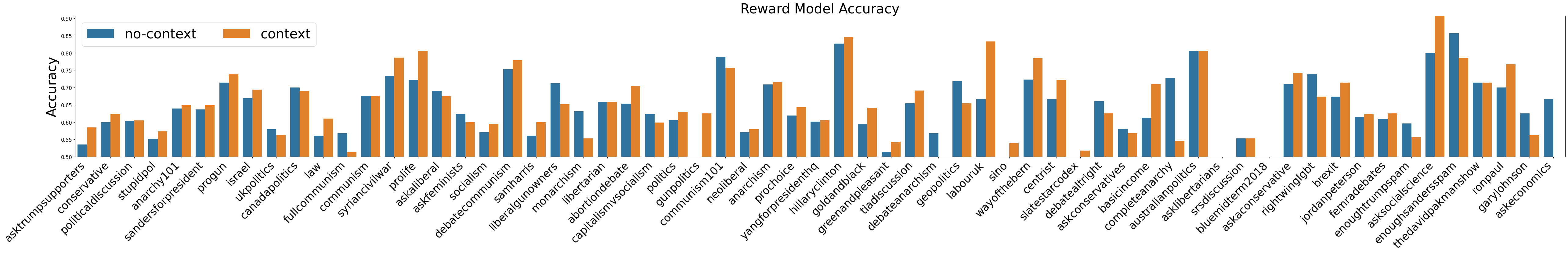}
    \caption{Subreddit Specific Preference Accuracy for Politics}
    \label{fig:politics-reward}
\end{figure*}

\begin{figure*}
    \centering
    \includegraphics[width=\textwidth]{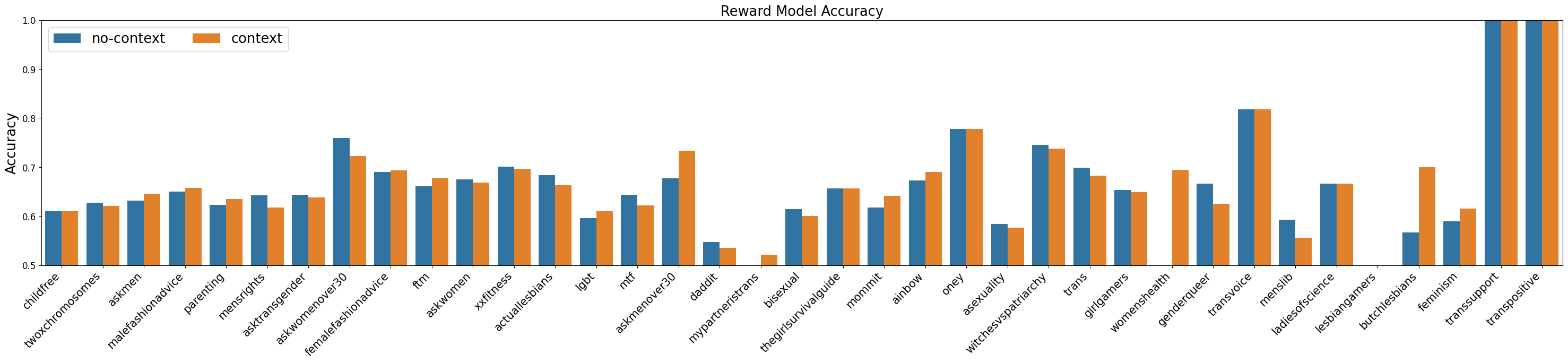}
    \caption{Subreddit Specific Preference Accuracy for Gender / Sexuality}
    \label{fig:gender-sexuality-reward}
\end{figure*}

We detail the subreddit-specific reward model (preference) accuracies in figures~\ref{fig:finance-reward}, \ref{fig:history-reward}, \ref{fig:science-reward}, \ref{fig:politics-reward} and \ref{fig:gender-sexuality-reward}.

\subsection{Subreddit specific win rates as judged by GPT-4}
\label{appsec:subreddit-win-rates}
\begin{figure*}
    \centering
    \includegraphics[width=\textwidth]{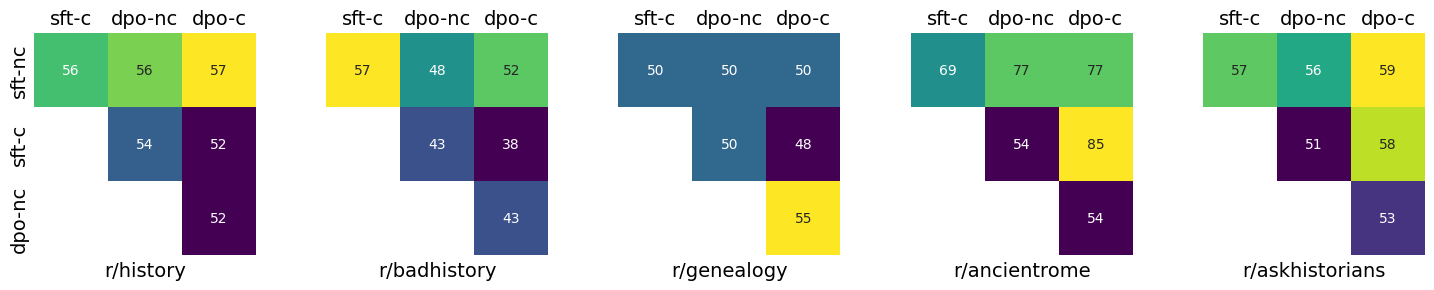}
    \caption{Subreddit-wise win-rates for History as judged by GPT-4}
    \label{fig:history-win-rates}
\end{figure*}

\begin{figure*}
    \centering
    \includegraphics[width=\textwidth]{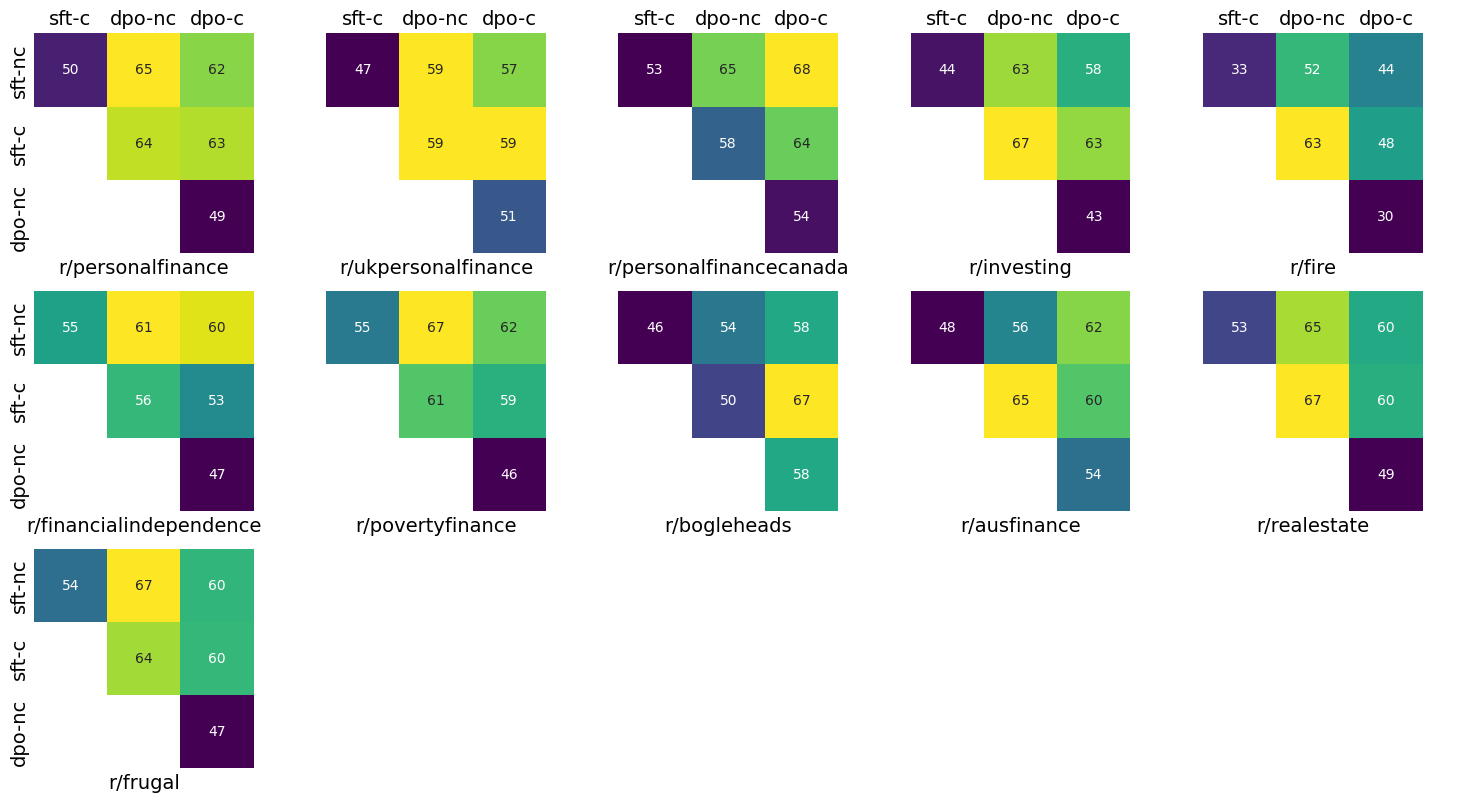}
    \caption{Subreddit-wise win-rates for Finance as judged by GPT-4}
    \label{fig:finance-win-rates}
\end{figure*}

\begin{figure*}
    \centering
    \includegraphics[width=\textwidth]{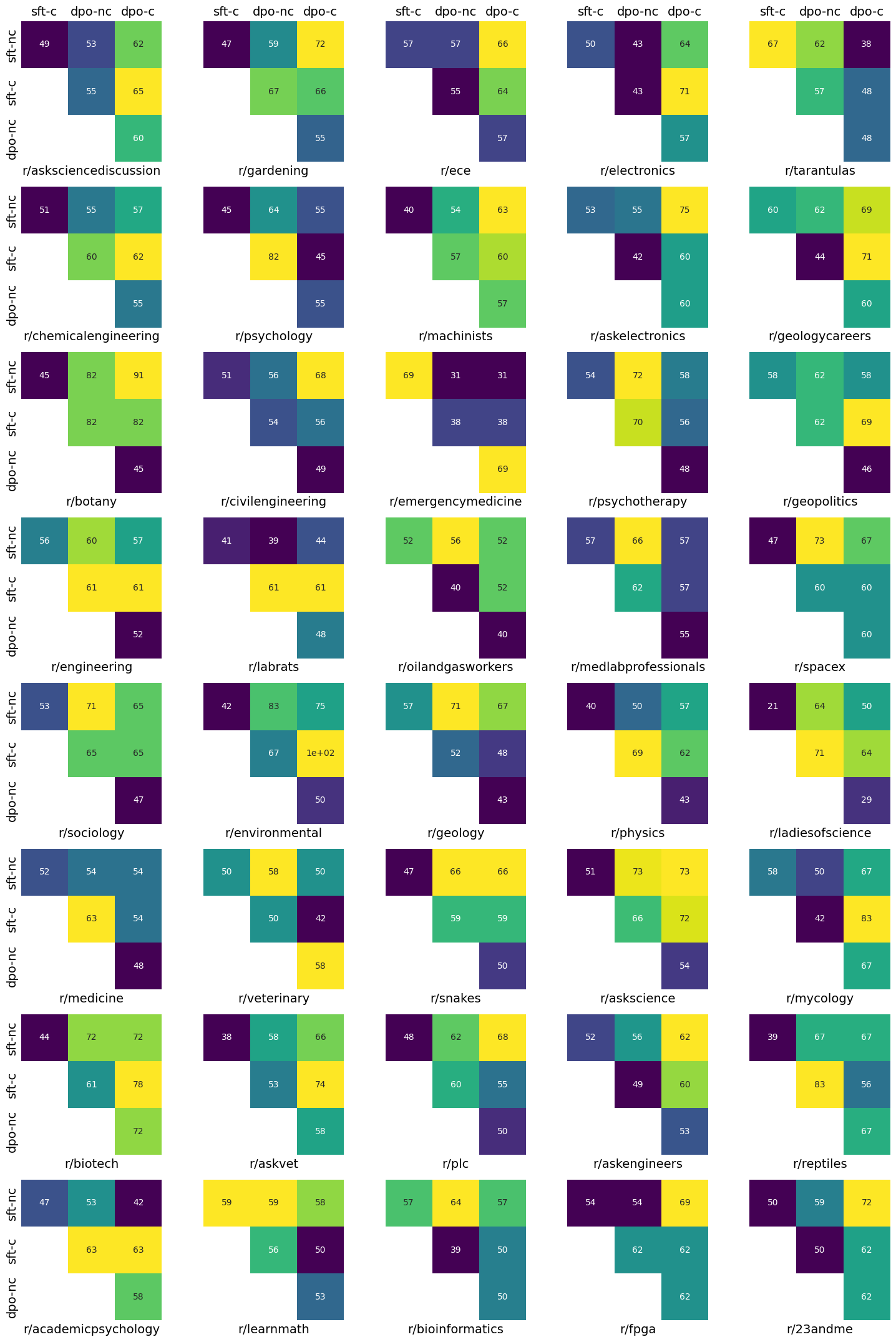}
    \caption{Subreddit-wise win-rates for Science as judged by GPT-4 (1/2)}
    \label{fig:science-win-rates-1}
\end{figure*}

\begin{figure*}
    \centering
    \includegraphics[width=\textwidth]{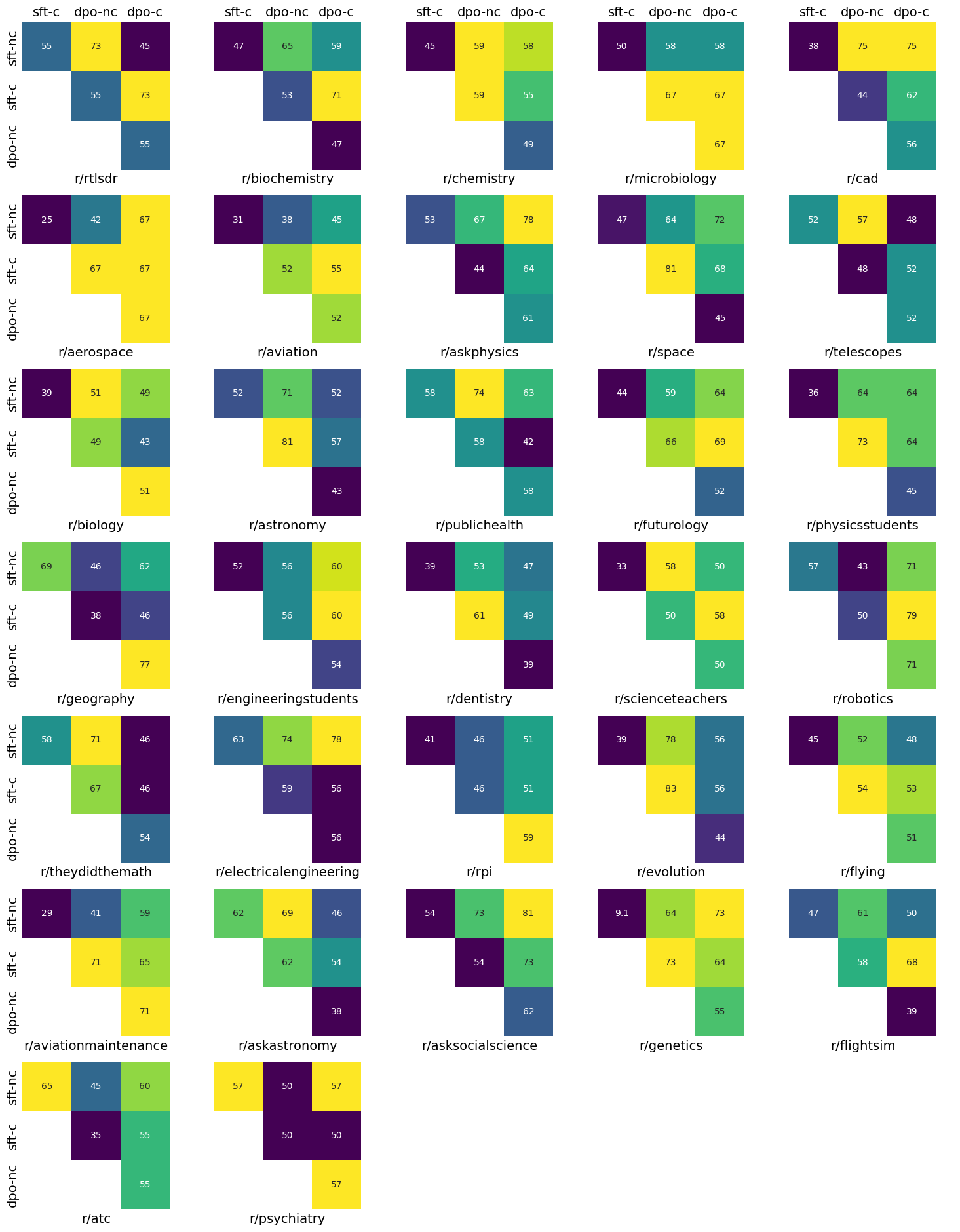}
    \caption{Subreddit-wise win-rates for Science as judged by GPT-4}
    \label{fig:science-win-rates-2}
\end{figure*}

\begin{figure*}
    \centering
    \includegraphics[width=\textwidth]{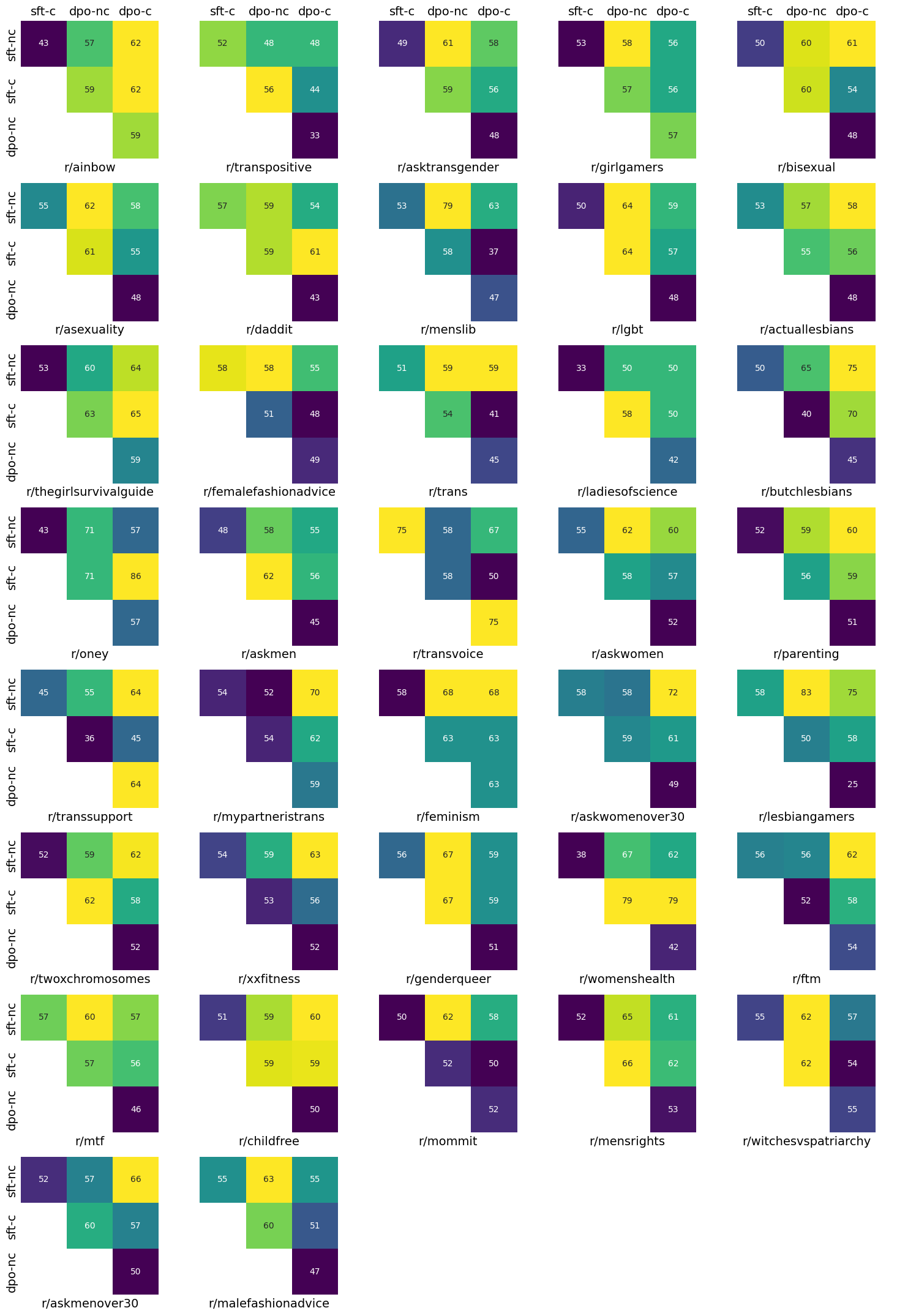}
    \caption{Subreddit-wise win-rates for Gender / Sexuality as judged by GPT-4}
    \label{fig:gender-win-rates}
\end{figure*}

\begin{figure*}
    \centering
    \includegraphics[width=\textwidth]{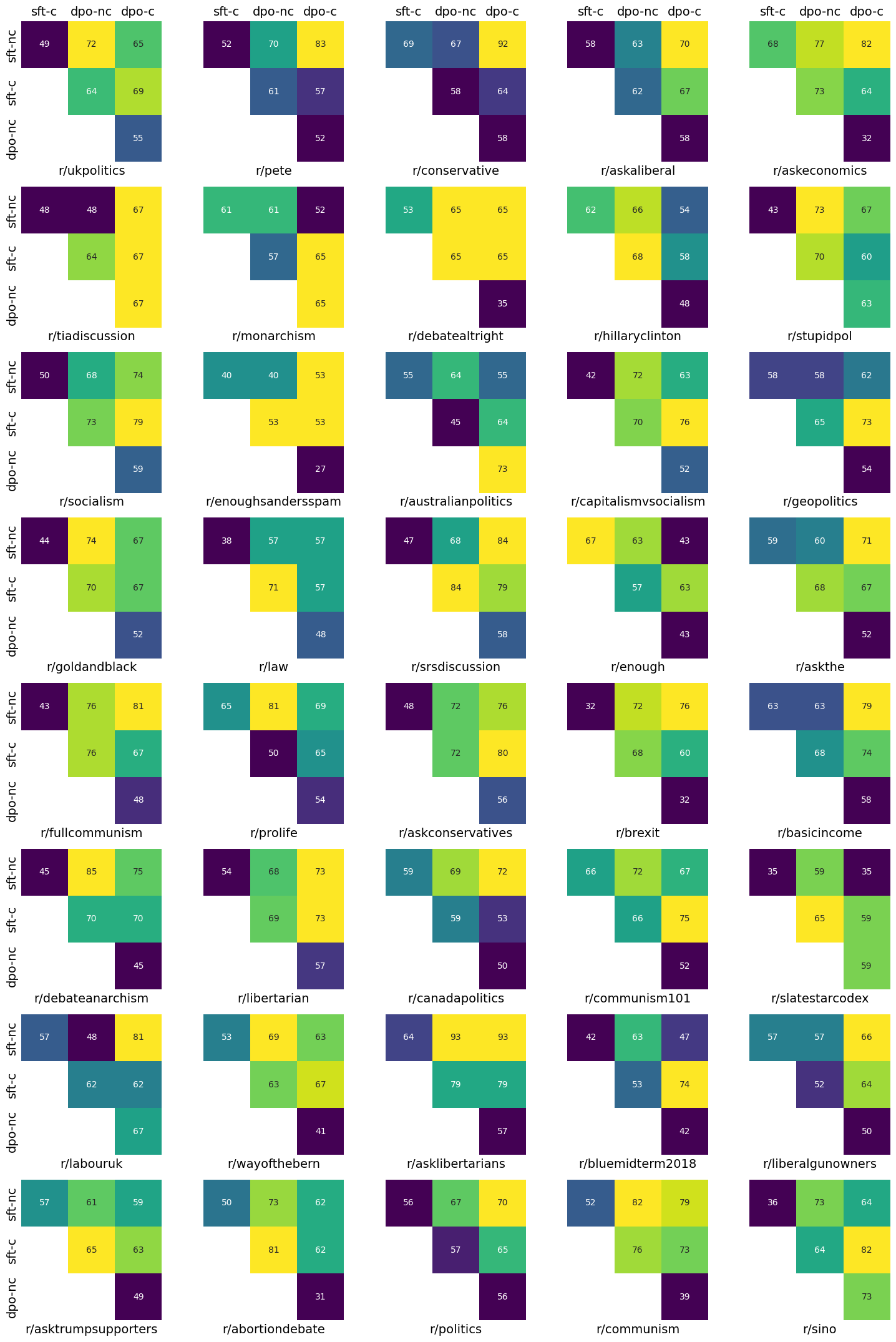}
    \caption{Subreddit-wise win-rates for Politics as judged by GPT-4 (1/2)}
    \label{fig:politics-win-rates-1}
\end{figure*}

\begin{figure*}
    \centering
    \includegraphics[width=\textwidth]{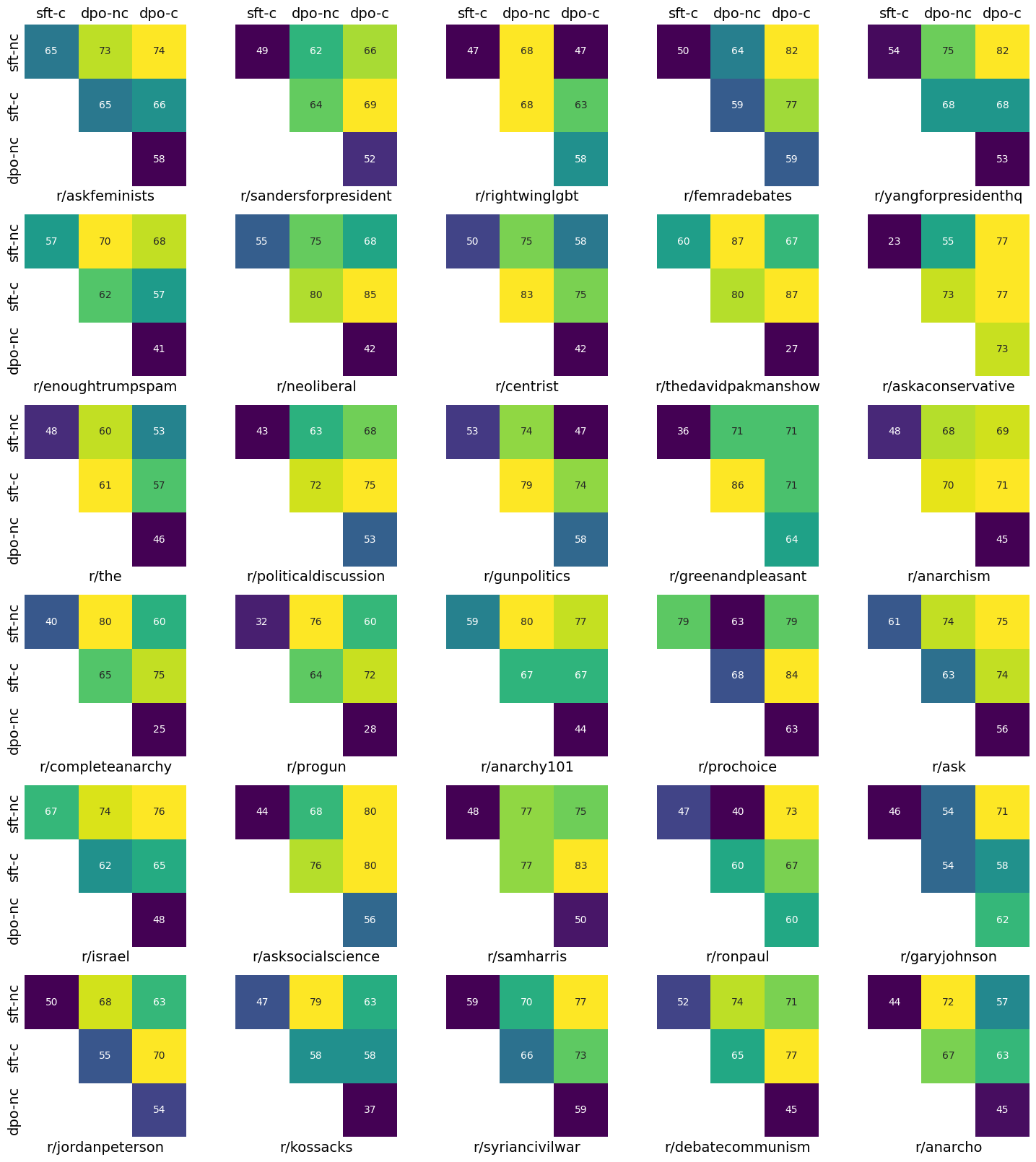}
    \caption{Subreddit-wise win-rates for Politics as judged by GPT-4 (2/2)}
    \label{fig:politics-win-rates-2}
\end{figure*}

We detail the subreddit-specific win-rates as judged by GPT-4 in figures~\ref{fig:finance-win-rates}, \ref{fig:history-win-rates}, \ref{fig:science-win-rates-1}, \ref{fig:science-win-rates-2} \ref{fig:politics-win-rates-1}, \ref{fig:politics-win-rates-2}, and
\ref{fig:gender-win-rates}.

\subsection{Subreddit Prediction Performance}
\label{appsec:subreddit-prediction-accuracies}
We detail the subreddit-predictability results in~\ref{table:finance_subreddit_prediction}, \ref{table:history_subreddit_prediction}, \ref{table:science_subreddit_prediction-1}, \ref{table:science_subreddit_prediction-2}, \ref{table:politics_subreddit_prediction-1}, \ref{table:politics_subreddit_prediction-2}, and
\ref{table:gender_sexuality_subreddit_prediction}. Subreddits with high prediction performance weakly correlate with lower win-rates.

\begin{table*}[h!]
\centering
\begin{tabular}{lrrrr}
\toprule
\textbf{Domain} & \textbf{Examples} & \textbf{Precision} & \textbf{Recall} & \textbf{F1} \\ \midrule
r/personalfinance & 1531 & 88.44 & 80.64 & 84.36 \\
r/frugal & 868 & 87.44 & 91.12 & 89.24 \\
r/povertyfinance & 190 & 51.58 & 70.50 & 59.57 \\
r/realestate & 354 & 85.59 & 88.34 & 86.94 \\
r/personalfinancecanada & 564 & 81.38 & 87.10 & 84.14 \\
r/financialindependence & 313 & 84.98 & 81.10 & 83.00 \\
r/investing & 455 & 89.45 & 86.97 & 88.19 \\
r/fire & 50 & 6.00 & 50.00 & 10.71 \\
r/ukpersonalfinance & 326 & 93.56 & 89.18 & 91.32 \\
r/ausfinance & 238 & 83.61 & 87.67 & 85.59 \\
r/bogleheads & 40 & 40.00 & 43.24 & 41.56 \\
\bottomrule
\end{tabular}
\caption{Finance Subreddit Prediction Performance}
\label{table:finance_subreddit_prediction}
\end{table*}

\begin{table*}[h!]
\centering
\begin{tabular}{lrrrr}
\toprule
\textbf{Domain} & \textbf{Examples} & \textbf{Precision} & \textbf{Recall} & \textbf{F1} \\ 
\midrule
r/history & 322 & 84.47 & 63.85 & 72.73 \\
r/askhistorians & 362 & 56.63 & 84.36 & 67.77 \\
r/genealogy & 156 & 98.72 & 97.47 & 98.09 \\
r/badhistory & 50 & 80.00 & 86.96 & 83.33 \\
r/ancientrome & 19 & 89.47 & 65.38 & 75.56 \\
\bottomrule
\end{tabular}
\caption{History Subreddit Prediction Performance}
\label{table:history_subreddit_prediction}
\end{table*}

\begin{table*}[h!]
\centering
\begin{tabular}{lrrrr}
\toprule
\textbf{Domain} & \textbf{Examples} & \textbf{Precision} & \textbf{Recall} & \textbf{F1} \\ 
\midrule
r/askscience & 435 & 83.45 & 71.88 & 77.23 \\
r/askelectronics & 74 & 81.08 & 68.97 & 74.53 \\
r/asksciencediscussion & 75 & 10.67 & 25.81 & 15.09 \\
r/theydidthemath & 35 & 100.00 & 94.59 & 97.22 \\
r/flightsim & 52 & 84.62 & 100.00 & 91.67 \\
r/biology & 94 & 70.21 & 65.35 & 67.69 \\
r/medicine & 223 & 91.48 & 90.67 & 91.07 \\
r/physics & 73 & 73.97 & 75.00 & 74.48 \\
r/oilandgasworkers & 50 & 82.00 & 89.13 & 85.42 \\
r/engineeringstudents & 321 & 72.27 & 70.52 & 71.38 \\
r/labrats & 88 & 67.05 & 61.46 & 64.13 \\
r/psychotherapy & 100 & 95.00 & 95.96 & 95.48 \\
r/learnmath & 98 & 87.76 & 83.50 & 85.57 \\
r/geopolitics & 41 & 100.00 & 91.11 & 95.35 \\
r/engineering & 136 & 19.85 & 34.62 & 25.23 \\
r/microbiology & 18 & 61.11 & 78.57 & 68.75 \\
r/rpi & 64 & 81.25 & 94.55 & 87.39 \\
r/flying & 499 & 95.59 & 92.98 & 94.27 \\
r/gardening & 107 & 96.26 & 98.10 & 97.17 \\
r/medlabprofessionals & 88 & 90.91 & 91.95 & 91.43 \\
r/space & 94 & 53.19 & 62.50 & 57.47 \\
r/genetics & 15 & 80.00 & 75.00 & 77.42 \\
r/evolution & 24 & 45.83 & 42.31 & 44.00 \\
r/plc & 63 & 96.83 & 89.71 & 93.13 \\
r/reptiles & 25 & 92.00 & 76.67 & 83.64 \\
r/snakes & 47 & 78.72 & 94.87 & 86.05 \\
r/futurology & 108 & 84.26 & 81.25 & 82.73 \\
r/bioinformatics & 39 & 92.31 & 87.80 & 90.00 \\
r/askengineers & 241 & 56.43 & 45.64 & 50.46 \\
r/askphysics & 45 & 17.78 & 44.44 & 25.40 \\
r/civilengineering & 74 & 86.49 & 59.26 & 70.33 \\
r/tarantulas & 31 & 83.87 & 96.30 & 89.66 \\
r/chemistry & 121 & 81.82 & 77.34 & 79.52 \\
r/chemicalengineering & 69 & 72.46 & 78.12 & 75.19 \\
r/ece & 83 & 33.73 & 43.75 & 38.10 \\
r/asksocialscience & 39 & 30.77 & 40.00 & 34.78 \\
r/atc & 40 & 77.50 & 75.61 & 76.54 \\
r/electronics & 22 & 9.09 & 25.00 & 13.33 \\
r/robotics & 19 & 100.00 & 100.00 & 100.00 \\
r/dentistry & 96 & 100.00 & 100.00 & 100.00 \\
r/biochemistry & 24 & 91.67 & 100.00 & 95.65 \\
r/veterinary & 22 & 77.27 & 80.95 & 79.07 \\
r/emergencymedicine & 22 & 54.55 & 80.00 & 64.86 \\
r/physicsstudents & 17 & 94.12 & 88.89 & 91.43 \\
r/geography & 26 & 80.77 & 77.78 & 79.25 \\
r/biotech & 27 & 62.96 & 77.27 & 69.39 \\
r/machinists & 60 & 95.00 & 86.36 & 90.48 \\
r/electricalengineering & 49 & 14.29 & 31.82 & 19.72 \\
\bottomrule
\end{tabular}
\caption{Science Subreddit Prediction Performance (2/2)}
\label{table:science_subreddit_prediction-1}
\end{table*}

\begin{table*}[h!]
\centering
\begin{tabular}{lrrrr}
\toprule
\textbf{Domain} & \textbf{Examples} & \textbf{Precision} & \textbf{Recall} & \textbf{F1} \\ 
\midrule
r/askvet & 72 & 87.50 & 98.44 & 92.65 \\
r/23andme & 53 & 94.34 & 98.04 & 96.15 \\
r/cad & 20 & 75.00 & 68.18 & 71.43 \\
r/aviation & 42 & 42.86 & 72.00 & 53.73 \\
r/academicpsychology & 25 & 84.00 & 65.62 & 73.68 \\
r/publichealth & 24 & 54.17 & 68.42 & 60.47 \\
r/geology & 28 & 32.14 & 40.91 & 36.00 \\
r/askastronomy & 16 & 43.75 & 41.18 & 42.42 \\
r/fpga & 17 & 0.00 & 0.00 & 0.00 \\
r/astronomy & 28 & 96.43 & 64.29 & 77.14 \\
r/aviationmaintenance & 25 & 72.00 & 69.23 & 70.59 \\
r/psychiatry & 26 & 65.38 & 73.91 & 69.39 \\
r/geologycareers & 77 & 76.62 & 79.73 & 78.15 \\
r/spacex & 32 & 93.75 & 85.71 & 89.55 \\
r/telescopes & 28 & 60.71 & 80.95 & 69.39 \\
r/psychology & 16 & 37.50 & 85.71 & 52.17 \\
r/scienceteachers & 16 & 0.00 & 0.00 & 0.00 \\
r/ladiesofscience & 23 & 30.43 & 31.82 & 31.11 \\
r/aerospace & 21 & 66.67 & 93.33 & 77.78 \\
r/mycology & 15 & 73.33 & 73.33 & 73.33 \\
r/botany & 15 & 33.33 & 83.33 & 47.62 \\
r/sociology & 26 & 76.92 & 95.24 & 85.11 \\
r/rtlsdr & 13 & 69.23 & 75.00 & 72.00 \\
\bottomrule
\end{tabular}
\caption{Science Subreddit Prediction Performance (2/2)}
\label{table:science_subreddit_prediction-2}
\end{table*}

\begin{table*}[h!]
\centering
\begin{tabular}{lrrrr}
\toprule
\textbf{Domain} & \textbf{Examples} & \textbf{Precision} & \textbf{Recall} & \textbf{F1} \\ 
\midrule
r/ukpolitics & 244 & 82.38 & 81.05 & 81.71 \\
r/politics & 131 & 58.78 & 67.54 & 62.86 \\
r/bluemidterm2018 & 29 & 72.41 & 65.62 & 68.85 \\
r/hillaryclinton & 111 & 78.38 & 84.47 & 81.31 \\
r/politicaldiscussion & 410 & 80.24 & 66.46 & 72.71 \\
r/askfeminists & 105 & 85.71 & 78.95 & 82.19 \\
r/capitalismvsocialism & 239 & 79.50 & 81.20 & 80.34 \\
r/anarchism & 190 & 60.00 & 67.06 & 63.33 \\
r/sandersforpresident & 395 & 90.13 & 86.62 & 88.34 \\
r/jordanpeterson & 141 & 28.37 & 53.33 & 37.04 \\
r/srsdiscussion & 37 & 70.27 & 66.67 & 68.42 \\
r/samharris & 112 & 81.25 & 83.49 & 82.35 \\
r/libertarian & 248 & 65.73 & 75.46 & 70.26 \\
r/liberalgunowners & 97 & 81.44 & 71.17 & 75.96 \\
r/asktrumpsupporters & 148 & 58.11 & 58.90 & 58.50 \\
r/stupidpol & 68 & 42.65 & 54.72 & 47.93 \\
r/debatealtright & 66 & 59.09 & 78.00 & 67.24 \\
r/askaliberal & 136 & 47.06 & 40.51 & 43.54 \\
r/abortiondebate & 80 & 91.25 & 71.57 & 80.22 \\
r/yangforpresidenthq & 243 & 94.65 & 96.64 & 95.63 \\
r/wayofthebern & 96 & 55.21 & 76.81 & 64.24 \\
r/askconservatives & 63 & 69.84 & 88.00 & 77.88 \\
r/communism & 51 & 66.67 & 64.15 & 65.38 \\
r/goldandblack & 39 & 10.26 & 57.14 & 17.39 \\
r/brexit & 49 & 51.02 & 73.53 & 60.24 \\
r/gunpolitics & 31 & 25.81 & 29.63 & 27.59 \\
r/progun & 48 & 37.50 & 36.73 & 37.11 \\
r/canadapolitics & 68 & 82.35 & 91.80 & 86.82 \\
r/basicincome & 32 & 37.50 & 70.59 & 48.98 \\
r/greenandpleasant & 41 & 78.05 & 91.43 & 84.21 \\
r/enoughsandersspam & 28 & 21.43 & 75.00 & 33.33 \\
r/debatecommunism & 57 & 80.70 & 83.64 & 82.14 \\
\bottomrule
\end{tabular}
\caption{Politics Subreddit Prediction Performance (1/2)}
\label{table:politics_subreddit_prediction-1}
\end{table*}

\begin{table*}[h!]
\centering
\begin{tabular}{lrrrr}
\toprule
\textbf{Domain} & \textbf{Examples} & \textbf{Precision} & \textbf{Recall} & \textbf{F1} \\ 
\midrule
r/law & 33 & 84.85 & 93.33 & 88.89 \\
r/geopolitics & 51 & 31.37 & 53.33 & 39.51 \\
r/askaconservative & 45 & 24.44 & 55.00 & 33.85 \\
r/labouruk & 40 & 55.00 & 81.48 & 65.67 \\
r/conservative & 64 & 20.31 & 68.42 & 31.33 \\
r/enoughtrumpspam & 58 & 56.90 & 91.67 & 70.21 \\
r/monarchism & 37 & 78.38 & 100.00 & 87.88 \\
r/socialism & 157 & 64.97 & 59.65 & 62.20 \\
r/israel & 109 & 88.07 & 93.20 & 90.57 \\
r/communism101 & 89 & 42.70 & 59.38 & 49.67 \\
r/askeconomics & 27 & 66.67 & 66.67 & 66.67 \\
r/anarchy101 & 103 & 55.34 & 55.88 & 55.61 \\
r/asksocialscience & 35 & 37.14 & 76.47 & 50.00 \\
r/syriancivilwar & 67 & 91.04 & 87.14 & 89.05 \\
r/slatestarcodex & 26 & 80.77 & 77.78 & 79.25 \\
r/neoliberal & 77 & 31.17 & 64.86 & 42.11 \\
r/tiadiscussion & 57 & 70.18 & 74.07 & 72.07 \\
r/centrist & 23 & 60.87 & 66.67 & 63.64 \\
r/asklibertarians & 31 & 58.06 & 78.26 & 66.67 \\
r/prolife & 63 & 31.75 & 35.71 & 33.61 \\
r/prochoice & 43 & 67.44 & 78.38 & 72.50 \\
r/debateanarchism & 43 & 25.58 & 39.29 & 30.99 \\
r/rightwinglgbt & 40 & 52.50 & 51.22 & 51.85 \\
r/fullcommunism & 32 & 87.50 & 84.85 & 86.15 \\
r/sino & 21 & 66.67 & 100.00 & 80.00 \\
r/femradebates & 52 & 42.31 & 66.67 & 51.76 \\
r/ronpaul & 23 & 65.22 & 65.22 & 65.22 \\
r/thedavidpakmanshow & 19 & 31.58 & 75.00 & 44.44 \\
r/garyjohnson & 36 & 77.78 & 80.00 & 78.87 \\
r/completeanarchy & 32 & 9.38 & 21.43 & 13.04 \\
r/australianpolitics & 20 & 65.00 & 72.22 & 68.42 \\
\bottomrule
\end{tabular}
\caption{Politics Subreddit Prediction Performance (2/2)}
\label{table:politics_subreddit_prediction-2}
\end{table*}

\begin{table*}[h!]
\centering
\begin{tabular}{lrrrr}
\toprule
\textbf{Domain} & \textbf{Examples} & \textbf{Precision} & \textbf{Recall} & \textbf{F1} \\ 
\midrule
r/childfree & 2412 & 97.01 & 95.16 & 96.08 \\
r/malefashionadvice & 477 & 91.40 & 92.37 & 91.89 \\
r/askmen & 1839 & 83.36 & 76.80 & 79.95 \\
r/twoxchromosomes & 1270 & 72.68 & 69.29 & 70.95 \\
r/askwomen & 1352 & 75.15 & 69.16 & 72.03 \\
r/xxfitness & 788 & 94.54 & 93.12 & 93.83 \\
r/parenting & 1530 & 92.09 & 82.83 & 87.22 \\
r/actuallesbians & 951 & 80.44 & 75.89 & 78.10 \\
r/ftm & 716 & 73.04 & 77.48 & 75.20 \\
r/daddit & 263 & 45.63 & 75.47 & 56.87 \\
r/askmenover30 & 219 & 35.62 & 50.65 & 41.82 \\
r/bisexual & 463 & 79.05 & 79.22 & 79.14 \\
r/femalefashionadvice & 618 & 91.75 & 86.30 & 88.94 \\
r/girlgamers & 210 & 99.05 & 87.39 & 92.86 \\
r/asktransgender & 1343 & 82.20 & 65.40 & 72.85 \\
r/thegirlsurvivalguide & 311 & 41.80 & 53.06 & 46.76 \\
r/asexuality & 193 & 89.64 & 96.11 & 92.76 \\
r/mensrights & 286 & 85.66 & 83.90 & 84.78 \\
r/feminism & 39 & 5.13 & 22.22 & 8.33 \\
r/mtf & 559 & 44.36 & 50.20 & 47.10 \\
r/ainbow & 71 & 0.00 & 0.00 & 0.00 \\
r/lgbt & 382 & 50.52 & 59.75 & 54.75 \\
r/menslib & 34 & 23.53 & 47.06 & 31.37 \\
r/oney & 31 & 3.23 & 100.00 & 6.25 \\
r/genderqueer & 56 & 33.93 & 65.52 & 44.71 \\
r/mommit & 270 & 47.04 & 68.65 & 55.82 \\
r/mypartneristrans & 95 & 74.74 & 83.53 & 78.89 \\
r/askwomenover30 & 255 & 39.22 & 60.61 & 47.62 \\
r/witchesvspatriarchy & 99 & 63.64 & 94.03 & 75.90 \\
r/trans & 120 & 0.00 & 0.00 & 0.00 \\
r/butchlesbians & 34 & 52.94 & 81.82 & 64.29 \\
r/transsupport & 13 & 0.00 & 0.00 & 0.00 \\
r/transvoice & 19 & 5.26 & 100.00 & 10.00 \\
r/lesbiangamers & 16 & 0.00 & 0.00 & 0.00 \\
r/transpositive & 33 & 0.00 & 0.00 & 0.00 \\
r/ladiesofscience & 15 & 80.00 & 92.31 & 85.71 \\
r/womenshealth & 40 & 25.00 & 47.62 & 32.79 \\ \bottomrule
\end{tabular}
\caption{Gender / Sexuality Subreddit Prediction Performance }
\label{table:gender_sexuality_subreddit_prediction}
\end{table*}
\end{document}